\documentclass{article}

\usepackage[utf8]{inputenc} 
\usepackage{microtype}
\usepackage{graphicx}
\usepackage{fontawesome}
\usepackage{booktabs} 

\usepackage{hyperref}

\usepackage[preprint]{icml2026}

\usepackage{amsmath}
\usepackage{amssymb}
\usepackage{mathtools}
\usepackage{amsthm}
\usepackage{amsfonts}
\usepackage{nicefrac}       

\usepackage{url}            
\usepackage{color,xcolor}   
\usepackage{caption}
\usepackage{subcaption}
\usepackage{setspace}
\usepackage{multirow}
\usepackage{threeparttable}
\usepackage{makecell}
\usepackage{pifont}
\usepackage{xspace}
\usepackage[misc]{ifsym}
\usepackage{colortbl}
\usepackage{wrapfig}

\newcolumntype{K}[1]{>{\centering\arraybackslash}p{#1}}

\hypersetup{
    pdftitle={},
    pdfsubject={},
    pdfauthor={},
    pdfkeywords={},
}

\usepackage[capitalize,noabbrev]{cleveref}

\theoremstyle{plain}

\theoremstyle{definition}

\theoremstyle{remark}

\usepackage[textsize=tiny]{todonotes}

\newcommand{\modelname}{\textsc{Seg-MoE}}
\newcommand{\model}{\modelname\xspace}

\newcommand{\smodel}{\modelname\textsubscript{small}\xspace}
\newcommand{\basemodel}{\modelname\textsubscript{base}\xspace}
\newcommand{\largemodel}{\modelname\textsubscript{large}\xspace}
\newcommand{\dmodel}{$d_{\text{model}}$\xspace}
\newcommand{\dff}{$d_{\text{ff}}$\xspace}

\newcommand{\bestres}[1]{{\textbf{\textcolor{red}{#1}}}}
\newcommand{\secondres}[1]{{\underline{\textcolor{blue}{#1}}}}

\definecolor{fbApp}{HTML}{ffe4e3}
\definecolor{tabhighlight}{HTML}{e5e5e5}

\def\Tabref#1{Table~\ref{#1}}

\def\eqref#1{equation~\ref{#1}}

\icmltitlerunning{\modelname: Multi-Resolution Segment-wise Mixture-of-Experts for Time Series Forecasting Transformers}

\begin{document}

\twocolumn[
\icmltitle{\modelname: Multi-Resolution Segment-wise Mixture-of-Experts for\\ Time Series Forecasting Transformers}



\icmlsetsymbol{equal}{*}

\begin{icmlauthorlist}

\icmlauthor{Evandro S. Ortigossa}{wis}
\icmlauthor{Eran Segal}{wis,mbzu}


\end{icmlauthorlist}


\icmlaffiliation{wis}{Department of Computer Science and Applied Mathematics, Weizmann Institute of Science, Rehovot, Israel}
\icmlaffiliation{mbzu}{Mohamed bin Zayed University of Artificial Intelligence, Abu Dhabi, UAE}

\icmlcorrespondingauthor{Eran Segal}{eran.segal@weizmann.ac.il}

\icmlkeywords{Machine Learning, ICML, Mixture-of-Experts, Forecasting, Transformers}

\vskip 0.3in
]



\printAffiliationsAndNotice{}  

\begin{abstract}
    Transformer-based models have recently made significant advances in accurate time-series forecasting, but even these architectures struggle to scale efficiently while capturing long-term temporal dynamics. Mixture-of-Experts (MoE) layers are a proven solution to scaling problems in natural language processing. However, existing MoE approaches for time-series forecasting rely on token-wise routing mechanisms, which may fail to exploit the natural locality and continuity of temporal data. In this work, we introduce Seg-MoE, a sparse MoE design that routes and processes contiguous time-step segments rather than making independent expert decisions. Token segments allow each expert to model intra-segment interactions directly, naturally aligning with inherent temporal patterns. We integrate Seg-MoE layers into a time-series Transformer and evaluate it on multiple multivariate long-term forecasting benchmarks. Seg-MoE consistently achieves state-of-the-art forecasting accuracy across almost all prediction horizons, outperforming both dense Transformers and prior token-wise MoE models. Comprehensive ablation studies confirm that segment-level routing is the key factor driving these gains. Our results show that aligning the MoE routing granularity with the inherent structure of time series provides a powerful, yet previously underexplored, inductive bias, opening new avenues for conditionally sparse architectures in sequential data modeling.
\end{abstract}

\section{Introduction}
\label{sec:intro}

    Significant advances in deep learning have enabled remarkable predictive performance in different modalities, including natural language and vision~\cite{scaling_laws}. Time series forecasting is a critical task across domains such as finance~\cite{nie2024survey}, energy management~\cite{ozcanli2020deep}, healthcare~\cite{zeevi2015personalized}, and climate modeling~\cite{wu2023interpretable}, where accurate predictions inform decision-making. Deep learning solutions, particularly Transformer-based models, are promising tools for improving long-term forecast accuracy and scalability~\cite{kaplan2020scaling}. However, the sequential and multivariate nature of time-series data introduces unique challenges compared to language or vision contexts. Temporal dependencies can be highly heterogeneous, including local patterns (e.g., short-term fluctuations) along with global structures (e.g., long seasonal cycles). Additionally, multivariate time series increase computational and modeling complexity~\cite{shao2024exploring}, making it challenging to capture complex temporal patterns efficiently.

    Standard Transformers are dense architectures that struggle to scale to long sequences without incurring high computation and memory costs~\cite{nie2022time}. The attention mechanism has quadratic complexity with respect to sequence length, leading to suboptimal performance in long-term forecasting settings where models require extended horizons to model time-specific patterns~\cite{vaswani2017attention,nie2022time,liu2024timerxl}. Mixture-of-Experts ($\operatorname{MoE}$) architectures have emerged to address such scaling bottlenecks, enabling conditional computation in which only a sparse subset of model parameters is dynamically activated for each input token. $\operatorname{MoE}$ layers expand model capacity without a proportional increase in computation by routing each input through a subset of ``expert'' networks~\cite{shazeer2017outrageously,fedus2022switch}. This approach promotes specialization while keeping computational costs comparable to those of smaller, dense models~\cite{fedus2022switch,gshard}. However, $\operatorname{MoE}$ also introduces challenges, such as expert imbalance, which is typically addressed with gating losses to stabilize training.
    
    The $\operatorname{MoE}$ application to time-series forecasting is still in its early stages~\cite{shi2024time,liu2024moirai}, and prior approaches have inherited the limitations of per-token routing from language models. In standard $\operatorname{MoE}$ Transformers, each time step (input token) is routed independently to experts~\cite{dai2024deepseekmoe}, ignoring inherent temporal contiguities in time-series data. This point-wise gating strategy is computationally efficient, but it can limit the exploration of local dependencies. For example, consecutive observations that together encode a local trend or seasonal event may be split and routed to different experts, thereby preventing any single expert from effectively modeling that pattern. In long-term forecasting, capturing both short-term patterns and long-term structures is critical, and a lack of local coherence in routing can reduce performance. Recent efforts in time-series modeling have scaled MoE-based Transformers to billion-parameter models, achieving state-of-the-art accuracy by leveraging sparse activation and pre-training on massive datasets~\cite{shi2024time,liu2024moirai}. However, these models still rely on token-level gating, which limits expert specialization to isolated time steps and underutilizes the structured, segment-wise nature of temporal signals present in real-world sequences.
    
    To bridge this gap, we introduce \model, a novel $\operatorname{MoE}$ design for sequential data such as time series, where locality and temporal context are critical. \model introduces segment-wise routing instead of conventional token-wise routing, by reshaping the input sequence into non-overlapping contiguous segments and routing them as units in an order-preserving manner. \model treats segments as the basic routing units, enabling the model to capture intra-segment interactions and local patterns that would be fragmented under token-wise routing. This design draws on the inductive bias that time-series patterns are often local and compositional \cite{wu2021autoformer}. As a result, experts can specialize in domain-specific patterns, such as cycles or volatility clusters. We demonstrate that \model elevates a Transformer-based forecaster to state-of-the-art performance on long-term benchmarks, outperforming well-known forecasters. Moreover, our approach maintains efficiency comparable to standard $\operatorname{MoEs}$ while yielding robust learning through segment-wise context aggregation.
    The main contributions of this research are as follows:
    \begin{itemize}
        \item We propose \model\footnote{\faGithub~Available at \href{https://github.com/evortigosa/segmoe_forecast}{https://github.com/evortigosa/segmoe\_forecast}}, a sparse $\operatorname{MoE}$ architecture that shifts from token-wise to segment-wise routing and processing. \model fosters improved specialization for temporal data while preserving the efficiency benefits of sparse computation.
        
        \item Through comprehensive experiments and ablation studies on multivariate benchmarks, we demonstrate that segment-wise routing is an inductive bias that outperforms dense and standard token-wise $\operatorname{MoE}$ baselines for long-term forecasting.
        
        \item We investigate the scaling behavior of \model with respect to the segment length and the number of experts. We validate and provide empirical guidance on segment size to maximize performance.
    \end{itemize}

\section{Related Work}
\label{sec:related}

\subsection{Time Series Forecasting} 
\label{subsec:ts_dl}

    Time series forecasting has undergone several paradigm shifts over the past decades. Classical statistical methods, including the ARIMA family \cite{williams2001multivariate,vagropoulos2016comparison}, exponential smoothing \cite{de2011forecasting}, and state-space models such as structural time series \cite{koopman2000stamp}, dominated the field for a long time due to their interpretability and theoretical guarantees under stationarity assumptions. Although effective for univariate, short-horizon forecasting, these approaches scale poorly to high-dimensional multivariate series and have limited ability to capture complex nonlinear dynamics in data. 
    Recent and intensive research on deep learning has transformed the field \cite{ortigossa2025time}. 
    Recurrent architectures \cite{hochreiter1997long,chung2014empirical}, along with sequence-to-sequence variants \cite{sutskever2014sequence}, became the standard for multivariate forecasting by offering superior modeling of long-range temporal dependencies. Subsequent innovations introduced convolutional alternatives \cite{abbaspourazad2023large}, attention-augmented recurrent networks \cite{qin2017dual}, and hybrid models that combine recurrence with temporal convolution mechanisms \cite{salinas2017deepar} to better capture seasonality and local patterns. However, recurrent-based models still process data sequentially, which limits their efficiency and memory usage for long sequences.

    The introduction of the Transformer architecture \cite{vaswani2017attention} marked a turning point in the modeling of sequence data. By replacing recurrence with attention mechanisms, Transformers enable parallel training and direct modeling of arbitrary-range dependencies. Early adaptations for time series forecasting, such as Informer \cite{zhou2021informer}, Autoformer \cite{wu2021autoformer}, and FEDformer \cite{zhou2022fedformer}, addressed the quadratic complexity of standard attention by using sparse attention, series decompositions, and frequency-domain filtering, respectively. More recent designs, such as PatchTST \cite{nie2022time}, iTransformer \cite{liu2023itransformer}, and TimeXer \cite{wang2024timexer}, achieved state-of-the-art performance in long-term forecasting benchmarks by leveraging channel independence, patching strategies, and enriched context approaches. Despite these advances, scaling Transformers remains challenging due to the dense activation of all model parameters at every time step. This inefficiency becomes particularly acute in long-term forecasting settings, where models must maintain high capacity in extended input contexts while remaining efficient within real-world latency and memory constraints.

\subsection{Sparse Mixture-of-Experts (MoE) for Transformers}
\label{subsec:moe}

    The Mixture-of-Experts ($\operatorname{MoE}$) strategy, initially proposed by \citet{jacobs1991adaptive} and later popularized in deep learning by \citet{shazeer2017outrageously}, enables conditional computation where only a subset of parameters (from ``experts'') is activated for each input. When integrated into Transformer blocks, sparse $\operatorname{MoE}$ layers replace dense feed-forward networks ($\operatorname{FFNs}$), yielding capacity expansion with nearly constant computational cost at inference time. In an $\operatorname{MoE}$ layer, a learnable gating mechanism is trained to assign token embeddings to different experts, decomposing complex tasks into simpler subtasks and allowing each expert network to specialize on a subset of the input space.
    \citet{shazeer2017outrageously} demonstrated that sparsely-gated $\operatorname{MoE}$ models obey neural scaling laws \cite{zhou2022mixture,shi2024time}, increasing model capacity with minimal overhead and often outperforming dense models with equivalent active parameter count.
    Subsequent innovations have scaled up Transformers to trillions of parameters. Switch Transformers \cite{fedus2022switch} demonstrated that using a single expert per token ($\operatorname{Top-1}$ routing) can reduce inter-expert communication overhead and accelerate training. GShard \cite{gshard} introduced a flexible multi-expert $\operatorname{Top-K}$ routing that improves specialization and load balancing among experts. Mixtral \cite{mixtralmoe} refined the stability of $\operatorname{MoE}$ routing and improved expert utilization through auxiliary regularization that encourages balanced expert usage, while DeepSeekMoE \cite{dai2024deepseekmoe} presented globally shared experts that scale across devices, and MoHETS \cite{ortigossa2026mohets} explored experts with heterogeneous architectural biases to improve specialization. 
    
    Standard $\operatorname{MoEs}$ operate at token-level granularity, meaning that each token embedding is routed independently to one or more experts \citet{shazeer2017outrageously}. This point-wise routing, while computationally convenient in language models, implicitly assumes that mapping tokens independently is an optimized approximation. However, it can be suboptimal for time series data, where contiguous time steps often encode coherent local structures, such as cycles, abrupt trend shifts, or event-related spikes. 
    Only recently have researchers begun to explore $\operatorname{MoE}$ architectures for time-series forecasting. Moirai-MoE \cite{liu2024moirai} pioneered the introduction of $\operatorname{MoE}$ into large time series models. Similarly, Time-MoE \cite{shi2024time} scaled to billion-parameter forecasting models pre-trained on billions of time points. Notably, both Moirai-MoE and Time-MoE rely on standard token-wise routing for decoder-only Transformer architectures.
    To our knowledge, no prior work has systematically investigated segment-wise routing and processing in $\operatorname{MoE}$ Transformers, that is, routing contiguous time steps (segments) to the same expert.
    The intuition is that aligning the routing granularity with the natural temporal structures in the data could provide a stronger inductive bias for the experts.
    
    This work directly addresses that gap, demonstrating that aligning routing granularity with the intrinsic structure of time series yields significant gains in both forecast accuracy and expert specialization. \model thus extends $\operatorname{MoE}$ beyond language-centric designs by introducing a routing mechanism that preserves local structures.

\section{Methodology}
\label{sec:methodology}

    \textbf{Problem Formulation.} We focus on the long-term forecasting task of multivariate time series using Transformer-based models~\cite{vaswani2017attention}. 
    Consider a set of multivariate time series $\mathbf{X} \in \mathbb{R}^{D \times T}$ with $D$ variables (also called channels or variates) and $T$ time points. Each $\mathbf{x}_t = [x_t^1, x_t^2, \dots, x_t^D]^\top \in \mathbb{R}^D$ represents the observations of all $D$ variables at time step $t$. 
    Given a context (look-back) window of length $L$, the goal is to forecast the next $H$ time steps, i.e., the forecast horizon. 
    Thus, we train a forecasting model to predict the future sequence $\mathbf{\hat{X}}_{T+1:T+H} \in \mathbb{R}^{D \times H}$, conditioned on the historical window $\mathbf{X}_{T-L+1:T} \in \mathbb{R}^{D \times L}$. 
    In a Transformer-based model, each input sequence $\mathbf{x}_t$ is first processed through a learnable embedding module that projects it into a \dmodel-dimensional space:
    \begin{equation}
        \mathbf{z}^0_t = \operatorname{Embedding}(\mathbf{x}_t).
        \label{in_embedding}
    \end{equation}
    Next, the embedded representations are fed into the Transformer backbone, which consists of $B$ stacked Transformer blocks. Each block applies a multi-head self-attention followed by a position-wise feed-forward network, with residual connections and normalization layers as follows:
    \begin{align}
        \mathbf{h}^b_t & = \operatorname{ATT} \left( \operatorname{Norm} \left( \mathbf{z}^{b-1}_t \right) \right) + \mathbf{z}^{b-1}_t, \label{eq:attn} \\
        \mathbf{z}^b_t & = \operatorname{FFN} \left( \operatorname{Norm} \left( \mathbf{h}^b_t \right) \right) + \mathbf{h}^b_t, \label{eq:ffn}
    \end{align}
    where $\operatorname{ATT}(\,\cdot\,)$ is the self-attention operation, $\operatorname{FFN}(\,\cdot\,)$ is the feed-forward network, and $\operatorname{Norm}(\,\cdot\,)$ are normalization layers~\cite{vaswani2017attention}. This Transformer architecture has been used in the backbone of multiple forecasting models, providing powerful sequence modeling capacity for capturing temporal dependencies.

    \begin{figure*}[!ht]
        \centering
        \includegraphics[width=0.86\linewidth]{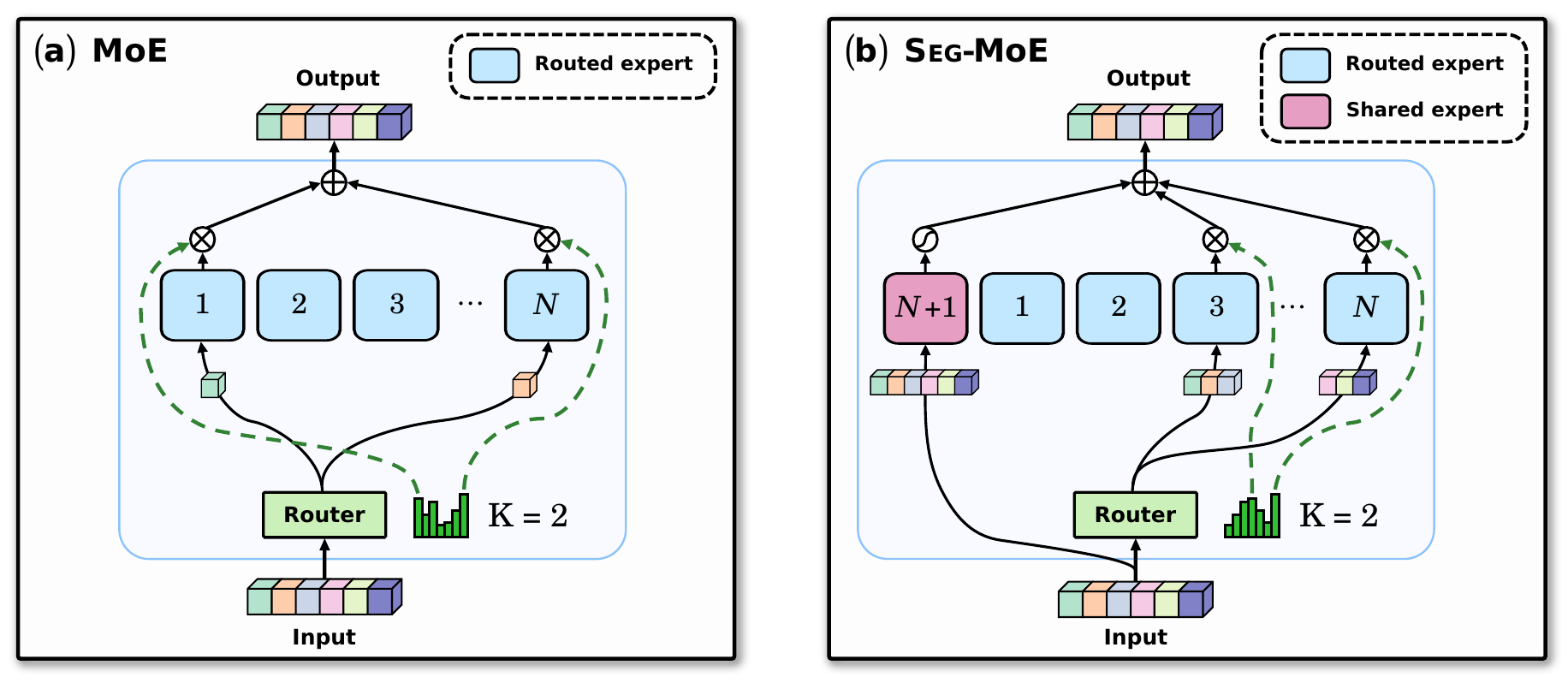}
        \caption{Mixture-of-Experts ($\operatorname{MoE}$) designs for sparse conditional computation in Transformer blocks. (a) Standard token-wise $\operatorname{MoE}$: a router computes token-to-expert affinities and selects $\operatorname{Top-K}$ routed experts from $N$ experts; the layer output is the weighted sum of the selected expert outputs. (b) \model: routing is performed at the segment level, and the output combines $\operatorname{Top-K}$ routed experts with an always-active shared expert, providing a stable, dense pathway while preserving sparsity in the routed experts.}
        \label{fig:smoe_arch}
    \end{figure*}

\subsection{Token-wise MoE Architecture}
\label{subsec:token_moe}

    In a standard Transformer, each input token (time step) is processed by dense layers, meaning that every token interacts with all the model's parameters. This dense computation becomes computationally expensive as the model size or the sequence length grows \cite{shi2024time}. Replacing $\operatorname{FFN}$ layers with $\operatorname{MoE}$ layers is a solution explored to introduce sparsity in Transformers (Figure~\ref{fig:smoe_arch}a). An $\operatorname{MoE}$ layer is composed of multiple parallel expert networks, each with the same architecture as a standard $\operatorname{FFN}$ \cite{shi2024time,dai2024deepseekmoe}.
    However, only a subset of these experts is activated for each token via a learned gating mechanism \cite{fedus2022switch,gshard}. 
    This design is based on conditional computation and selective activation, enabling parameter scaling while maintaining computational efficiency \cite{shazeer2017outrageously}, as each token uses only a fraction of the network. For a Transformer block $b$, the $\operatorname{FFN}$ is replaced by a $\operatorname{MoE}$ as follows:
    \begin{small}
    \begin{align}
        \operatorname{MoE} & ( \operatorname{Norm} ( \mathbf{h}^b_t ) ) = \sum_{i=1}^{N} ( {g_{i,t} \operatorname{FFN}_{i} ( \operatorname{Norm} ( \mathbf{h}^b_t ) )} ), \label{equ:mixture} \\
        g_{i,t} & = \begin{cases} 
        s_{i,t}, & s_{i,t} \in \operatorname{Top-K} (\{ s_{j, t} | 1 \leq j \leq N \}, K), \\
        0, & \text{otherwise}, 
        \end{cases} \label{equ:expert_score} \\
        s_{i,t} & = \operatorname{Softmax}_i ( \mathbf{W}_{i}^{b} ( \operatorname{Norm} ( \mathbf{h}^b_t ) ) ) \label{equ:expert_gate}, 
    \end{align}
    \end{small}
    
    where $N$ is the total number of experts in the $\operatorname{MoE}$ module, and $K$ is the number of experts activated for each token ($K \ll N$, typically $K=1$ or $2$). The gating function (Equation~\ref{equ:expert_gate}) computes a set of $N$ affinity scores $\{s_{1,t}, \dots, s_{N,t}\}$ for each token $t$, taking the $\operatorname{Softmax}$ logits from a learned projection $\mathbf{W}_{i}^{b} \in \mathbb{R}^{d_{\text{model}} \times N}$. Experts with the $\operatorname{Top-K}$ highest scores are selected (Equation~\ref{equ:expert_score}), and their corresponding outputs $\operatorname{FFN}{i}(\,\cdot\,)$ are weighted by $g{i,t}=s_{i,t}$ and combined to produce the $\operatorname{MoE}$ layer output (Equation~\ref{equ:mixture}) \cite{shi2024time,ortigossa2026mohets}. Therefore, each time point is processed by only $K$ out of $N$ experts, allowing different experts to specialize in different temporal patterns. This sparse architecture enables us to scale up a Transformer model capacity (by increasing $N$) without a proportional increase in computational cost, thereby improving efficiency and performance~\cite{liu2024moirai,dai2024deepseekmoe}.

\subsection{Transformer Backbone}
\label{subsec:overview}

    To experiment with \model, we use an encoder-only Transformer that leverages recent advances in large-scale time-series and language models. In particular, we embed the input sequences (Equation~\ref{in_embedding}) using a patching approach that converts the look-back window $\mathbf{X}_{T-L+1:T}$ into $M = \lceil L / P \rceil$ non-overlapping patch embeddings with length $P$~\cite{liu2024moirai,wang2024timexer}. To handle $n$-dimensional multivariate time series data, we adopt the channel-independence approach from~\cite{nie2022time}.
    
    We use $\operatorname{RMSNorm}$~\cite{zhang2019root} instead of the standard $\operatorname{LayerNorm}$~\cite{ba2016layer} at each Transformer sub-layer to normalize inputs and stabilize training (see Equations~\ref{eq:attn} and \ref{eq:ffn}). Moreover, we combine FlashAttention~\cite{dao2022flashattention} with grouped-query attention (GQA)~\cite{ainslie2023gqa} to increase efficiency and optimize memory usage of scaled dot-product computations. Also, we replace absolute positional encodings with Rotary Position Embeddings (RoPE) \cite{su2024roformer}, using the standard base frequency of $10,000$, because relative ordering is often more informative than absolute positioning in time-series data~\cite{erturk2025beyond}. Refer to Appendix~\ref{apdx:components} for implementation details.

\subsection{\model: The Segment-wise MoE Architecture}
\label{subsec:segmoe}

    We introduce conditional sparsity into time-series Transformers by replacing the standard feed-forward network ($\operatorname{FFN}$) in each Transformer block with \model, a segment-wise Mixture-of-Experts layer (Figure~\ref{fig:smoe_arch}b). \model is designed to exploit the temporal contiguity that characterizes time-series data. In contrast to token-wise $\operatorname{MoE}$ variants used in language models and recent time-series adaptations~\cite{shi2024time,liu2024moirai}, which route each time token independently, \model performs \textbf{routing at the segment level}. Contiguous, non-overlapping time-step segments share a single routing decision and are processed together by the selected experts. This design exposes segment-consistent interactions to expert transformation, an inductive bias particularly relevant for short-term trends spanning multiple adjacent patches.

    The core motivation is that many temporal patterns are local and compositional \cite{wu2021autoformer}. Expert assignment based on isolated tokens may ignore informative correlations across neighboring time steps, reducing the models' ability to capture semantically coherent motifs. In addition, isolated tokens may be dominated by noise or partial signals. By aligning routing granularity with temporal locality, \model encourages experts to specialize in coherent segment-level structures while retaining the computational advantages of sparse expert activation.
    
    \textbf{Segment Construction.} Let $\mathbf{H}^b \in \mathbb{R}^{M \times d_{\text{model}}}$ denote the normalized input sequence to the \model sub-layer at Transformer block $b$, where $M$ is the number of patch tokens (see Equation~\ref{equ:mixture}).
    Given a segment length $\omega_b$, we partition $\mathbf{H}^b$ into $C=\left\lceil{M}/{\omega_b} \right\rceil$ non-overlapping segments:
    \begin{equation}
        \mathbf{u}_c^b \in \mathbb{R}^{\omega_b \times d_{\text{model}}}, \quad c \in \{1,\dots,C\},
    \end{equation}
    where the final segment is right-padded with zeros if $M \bmod \omega_b \neq 0$, and a mask is used to ensure that the router input for the padded segment does not introduce bias.
    
    \textbf{Segment Routing.} Routing is computed for each segment with a lightweight linear gating network $\mathbf{W}^b \in \mathbb{R}^{(\omega_b \cdot d_{\text{model}})\times N}$, where $N$ is the number of routed experts. We flatten each segment to $\tilde{\mathbf{u}}_c^b=\mathrm{vec}(\mathbf{u}_c^b)\in\mathbb{R}^{\omega_b \cdot d_{\text{model}}}$ and forward to $\mathbf{W}^b$, allowing the routing gate to attend to intra-segment structures. A Softmax converts routing logits into probabilities $s_{i,c}$ as in Equation~\ref{equ:expert_gate}, and sparse activation is enforced with a $\operatorname{Top-K}$ selection (Equation~\ref{equ:expert_score}), activating only $K$ out of $N$ experts per segment~\cite{fedus2022switch}.

    \textbf{Shared Fallback Expert.} Following recent $\operatorname{MoE}$ designs~\cite{shi2024time,ortigossa2026mohets}, \model includes a shared expert that is applied to every segment, providing a stable, always-active pathway while routed experts specialize. To avoid architectural asymmetry, the shared expert operates on the same segmented representation as the routed experts. Formally, a \model layer combines one shared expert and \(N\) routed experts as follows:
    
    \begin{equation}
        \begin{aligned}
            \operatorname{S\mbox{\scriptsize{EG}}-MoE} ( \mathbf{{u}}^b_{c} ) & = g_{N+1,c} \operatorname{FFN}_{N+1} ( \mathbf{{u}}^b_{c} ) \\
            & + \sum_{i=1}^{N} ( {g_{i,c} \operatorname{FFN}_{i} ( \mathbf{{u}}^b_{c} )} ), \label{equ:segmoe}
        \end{aligned}
    \end{equation}
    
    where $g_{N + 1,c}$ is a $\operatorname{Sigmoid}$ gate that modulates the shared expert contribution~\cite{shi2024time}, and $g_{i,c}$ are the router gate weights (Equation~\ref{equ:expert_score}).
    
    \textbf{Integration.} \model is a modular layer component designed to replace dense $\operatorname{FFNs}$ or standard $\operatorname{MoEs}$ in Transformer blocks, preserving the residual structure of the backbone model while increasing capacity through conditional computation.
    
    \textbf{Multi-Resolution Design.} Time series often contain multi-scale structure (e.g., intra-day fluctuations and weekly trends). To capture this dynamic, we introduce multi-resolution routing by varying the segment length across Transformer blocks. Users may set a single scalar $\omega_b=\omega$ (uniform resolution) or provide a list of length $B$, with one scalar for each \model module (layer-wise multi-resolution). 
    When $\omega_b=1$, \model reduces to a standard token-wise $\operatorname{MoE}$ layer. Layer-wise multi-resolution introduces a controllable temporal hierarchy without dynamic routing overhead at inference time (segment sizes are fixed per layer), and empirically improves robustness to heterogeneous temporal dynamics. Thus, a multi-resolution \model is a natural extension to exploit the segment-wise paradigm, granting forecasters a temporal hierarchy that dense Transformers and uniform $\operatorname{MoEs}$ lack.

    Therefore, \model shifts token-wise routing towards segment-wise routing and segment-level expert processing, turning $\operatorname{MoE}$ layers from a generic scaling mechanism into a domain-aware inductive bias aligned with temporally contiguous modalities.
    Refer to Appendix~\ref{apdx:moe_components} for additional implementation details on the routing components.

\subsection{Loss Function}
\label{subsec:loss}

    Training large Transformer models with sparse $\operatorname{MoE}$ layers imposes stability challenges due to the large number of parameters, conditional computation embedded in near-discrete routing mechanisms, and the high variability of real-world data \cite{han2024fusemoe}. These models are typically trained using smooth convex functions such as Cross-Entropy or Mean Squared Error (MSE) \cite{nie2022time,wang2024timexer,dai2024deepseekmoe}. To improve robustness, we depart from standard losses and use the Huber loss \cite{huber1992robust,wen2019robusttrend} for the prediction task ($\mathcal{L}_{\text{pred}}$). The Huber loss combines the advantages of the L1 and MSE losses, making the model less sensitive to outliers and, thus, improving training stability \cite{shi2024time}. Further definitions are in Appendix~\ref{apdx:loss}. 
    
    However, the conditional computation derived from the selection mechanisms of $\operatorname{MoE}$ architectures renders their optimization inherently nonconvex \cite{li2022branch}. Optimizing only the prediction loss is insufficient and can degrade model capacity and training stability due to \textbf{load imbalance} and risk of routing collapse. Specifically, the model may learn to route virtually all token embeddings to only a few or even a single expert, leaving other experts underutilized and poorly trained \cite{dai2024deepseekmoe,shazeer2017outrageously}. To prevent this risk of collapse and encourage balanced expert utilization under sparse activation, we incorporate an \textbf{auxiliary routing-balance loss} that penalizes uneven expert usage. This auxiliary loss builds on prior work in sparse expert models \cite{gshard,fedus2022switch,liu2024moirai,shi2024time} and is formalized as:
    \begin{small}
    \begin{align}
        \mathcal{L}_{\text{aux}} = & N \sum_{i=1}^{N}f_i r_i, \quad r_i = \frac{1}{C} \sum_{c=1}^{C} s_{i,c}, \label{equ:aux_loss} \\
        f_i = \frac{1}{KC} \sum_{c=1}^{C} \mathbb{I} & \left(\text{Time segment } c \text{ selects Expert } i \right),
    \end{align}
    \end{small}
    \hspace{-3pt}where $f_i$ is the fraction of segments routed to expert $i$, $C$ is the total number of segments in the sequence batch, and $r_i$ is the average router probability assigned to expert $i$ (Equation~\ref{equ:expert_score}). $\mathbb{I}$ is an indicator function that equals 1 when segment $c$ is routed to expert $i$~\cite{dai2024deepseekmoe}. The balance loss is small when the expert usage $f_i$ matches its allocated probability $r_i$, thereby encouraging a more uniform traffic distribution of segments among all experts.

\subsection{Training Objective: Long-Term Forecasting}
\label{subsec:objective}

    The training objective is a weighted combination of the prediction and auxiliary balance losses, ensuring both forecast accuracy and balanced expert utilization. The final loss is defined as: 
    \begin{equation}
        \mathcal{L} = \mathcal{L}_{\text{pred}} \left(\mathbf{{X}}_{T+1:T+H_o}, \hat{\mathbf{X}}_{T+1:T+H_o} \right) + \alpha \mathcal{L}_{\text{aux}},\label{equ:loss}
    \end{equation}
    where $H_o$ is a hyperparameter that defines the output length (i.e., the number of future time points predicted in each autoregressive step), and $\alpha$ is a scaling factor that controls the influence of the auxiliary balance loss. $\alpha$ is often set to a small value to encourage expert utilization without overtaking the prediction loss \cite{dai2024deepseekmoe}.

    \textbf{Look-back window.} The patch length $P$ is a hyperparameter that is set uniformly across the model, from the input embedding layer to the output head. We empirically select the input window size to be $L=512$ time steps, which we found to be an optimized length for long-term forecasting in accordance with previous work~\cite{nie2022time}. 

    \textbf{One-for-all forecasting.} We adopt a one-for-all forecasting strategy, training a single model to produce predictions for arbitrary forecast horizons $H$~\cite{liu2024timerxl}. During inference, forecasting is performed autoregressively, i.e., the model predicts the next time points $H_o$ at each iteration, then appends and forwards the updated sequence until it reaches the forecast horizon $H$. In our experiments, we evaluate the performance of \model on standard forecast horizons $H \in \{96, 192, 336, 720\}$.

\subsection{Hyperparameter and Optimization Settings}
\label{subsec:settings}

    We implement \model in PyTorch~\cite{paszke2019pytorch} and conduct all experiments on a single NVIDIA A100 80GB Tensor Core GPU. We use bfloat16 (BF16) precision for training to optimize memory efficiency and throughput with minimal impact on accuracy~\cite{kalamkar2019study}. 
    
    \vspace{1pt}\textbf{Hyperparameter settings.} We experiment with a range of model sizes and hyperparameters to find a good balance between performance and efficiency. In particular, the number of Transformer blocks is searched over $B\in\{4, 6, 8\}$, the embedding dimension \dmodel over $\{128, 256\}$, the patch length over $P \in \{8, 16\}$, and the output length over $H_o \in \{16, 24, 32\}$ time steps.
    Previous work has shown optimized results by setting $\operatorname{MoE}$ layers with $N = 8$ experts and $K = 2$ for the $\operatorname{Top-K}$ expert selection mechanism~\cite{gshard,dai2024deepseekmoe,liu2024moirai,shi2024time}, a setting that offers a good trade-off between model capacity and computational efficiency in different contexts~\cite{shi2024time}. 
    As \model proposes a novel routing approach for $\operatorname{MoEs}$, we extend standard settings to ensure optimal results in the context of segment-wise routing. Thus, we experiment with $N \in \{4, 8\}$, $K \in \{1, 2\}$, and varying the segment resolution over the interval $\omega = [1,5]$ across \model layers.
    
    \vspace{1pt}\textbf{Optimization settings.} We perform training using the AdamW optimizer \cite{loshchilov2017decoupled}. The learning rate is searched from the interval $3.2\mathrm{e}\text{-}{4}$ to $1.2\mathrm{e}\text{-}{6}$, with a linear warm-up over the first $10\%$ of the training steps, decayed following a cosine annealing schedule \cite{loshchilov2016sgdr}. We set AdamW's $\beta_1 = 0.9$, $\beta_2 = 0.95$, and a relatively high weight decay of $10^{-1}$, which is a recommendation for training large Transformer models~\cite{chen2020generative,dosovitskiy2020image}. Additionally, we use early stopping with a patience of 5 epochs to halt training if the validation loss stops improving. We fix the Huber loss parameter $\delta = 2.0$ (Equation~\ref{equ:huber}) and set the auxiliary loss scaling factor $\alpha = 0.02$ (Equation~\ref{equ:loss}), which is in line with the findings of \citet{shi2024time}. We do not apply any data augmentation techniques and train for a minimum of $10$ and a maximum of $30$ epochs, depending on the batch size used for each dataset. Refer to Appendix~\ref{apdx:settings} for additional implementation details and settings.

\section{Main Results}
\label{sec:results}

    \begin{table*}[!ht]
        \centering
        \caption{Long-term multivariate forecasting experiments. A lower MSE or MAE indicates a better prediction. Full-shot results are obtained from~\cite{liu2023itransformer,wang2024timexer,han2024softs}. {\bestres{Bold red}}: is the best value, \secondres{underlined blue}: the second best. $1^{\text{st}}$ Count is the number of wins achieved by models across prediction lengths and datasets.}
        \label{tab:ltsf_full}
        \resizebox{\linewidth}{!}{
        \renewcommand{\tabcolsep}{3pt}
        \begin{tabular}{cr|cc|cc|cc|cc|cc|cc|cc|cc|cc|cc|cc}
        \toprule
            \multicolumn{2}{c}{\scalebox{1.1}{\textbf{Models}}} & \multicolumn{2}{c}{\textbf{{\model}}} & \multicolumn{2}{c}{\textbf{SOFTS}} & \multicolumn{2}{c}{\textbf{TimeXer}} & \multicolumn{2}{c}{\textbf{iTransformer}} & \multicolumn{2}{c}{\textbf{TimeMixer}} &  \multicolumn{2}{c}{\textbf{TimesNet}} & \multicolumn{2}{c}{\textbf{PatchTST}} & \multicolumn{2}{c}{\textbf{Crossformer}} & \multicolumn{2}{c}{\textbf{TiDE}} & \multicolumn{2}{c}{\textbf{DLinear}} & \multicolumn{2}{c}{\textbf{FEDformer}} \\
    
            \cmidrule(lr){3-4} \cmidrule(lr){5-6}\cmidrule(lr){7-8} \cmidrule(lr){9-10}\cmidrule(lr){11-12}\cmidrule(lr){13-14}\cmidrule(lr){15-16}\cmidrule(lr){17-18}\cmidrule(lr){19-20} \cmidrule(lr){21-22} \cmidrule(lr){23-24}

            \multicolumn{2}{c}{\scalebox{1.1}{\textbf{Metrics}}} & \textbf{MSE} & \textbf{MAE} & \textbf{MSE} & \textbf{MAE} & \textbf{MSE} & \textbf{MAE} & \textbf{MSE} & \textbf{MAE} & \textbf{MSE} & \textbf{MAE} & \textbf{MSE} & \textbf{MAE} & \textbf{MSE} & \textbf{MAE} & \textbf{MSE} & \textbf{MAE} & \textbf{MSE} & \textbf{MAE} & \textbf{MSE} & \textbf{MAE} & \textbf{MSE} & \textbf{MAE} \\
        \midrule
        \multirow{4}[1]{*}{ETTh1}
            & 96 & \bestres{0.343} & \bestres{0.381} & 0.381 & \secondres{0.399} & 0.382 & 0.403 & 0.386 & 0.405 & \secondres{0.375} & 0.400 & 0.384 & 0.402 & 0.414 & 0.419 & 0.423 & 0.448 & 0.479 & 0.464 & 0.386 & 0.400 & 0.376 & 0.419 \\
            & 192 & \bestres{0.378} & \bestres{0.405} & 0.435 & 0.431 & 0.429 & 0.435 & 0.441 & 0.436 & 0.436 & \secondres{0.429} & 0.421 & \secondres{0.429} & 0.460 & 0.445 & 0.471 & 0.474 & 0.525 & 0.492 & 0.437 & 0.432 & \secondres{0.420} & 0.448 \\
            & 336 & \bestres{0.394} & \bestres{0.419} & 0.480 & 0.452 & 0.468 & \secondres{0.448} & 0.487 & 0.458 & 0.484 & 0.458 & 0.491 & 0.469 & 0.501 & 0.466 & 0.570 & 0.546 & 0.565 & 0.515 & 0.481 & 0.459 & \secondres{0.459} & 0.465 \\
            & 720 & \bestres{0.408} & \bestres{0.441} & 0.499 & 0.488 & \secondres{0.469} & \secondres{0.461} & 0.503 & 0.491 & 0.498 & 0.482 & 0.521 & 0.500 & 0.500 & 0.488 & 0.653 & 0.621 & 0.594 & 0.558 & 0.519 & 0.516 & 0.506 & 0.507 \\
            \rowcolor{tabhighlight}
            & {\textbf{Avg.}} & \bestres{0.381} & \bestres{0.412} & 0.449 & 0.442 & \secondres{0.437} & \secondres{0.437} & 0.454 & 0.447 & 0.448 & 0.442 & 0.454 & 0.450 & 0.469 & 0.454 & 0.529 & 0.522 & 0.540 & 0.507 & 0.455 & 0.451 & 0.440 & 0.459 \\
        \midrule
        \multirow{4}[0]{*}{ETTh2}
            & 96 & \bestres{0.272} & \bestres{0.331} & 0.297 & 0.347 & \secondres{0.286} & \secondres{0.338} & 0.297 & 0.349 & 0.289 & 0.341 & 0.340 & 0.374 & 0.302 & 0.348 & 0.745 & 0.584 & 0.400 & 0.440 & 0.333 & 0.387 & 0.358 & 0.397 \\
            & 192 & \bestres{0.334} & \bestres{0.370} & 0.373 & 0.394 & \secondres{0.363} & \secondres{0.389} & 0.380 & 0.400 & 0.372 & 0.392 & 0.402 & 0.414 & 0.388 & 0.400 & 0.877 & 0.656 & 0.528 & 0.509 & 0.477 & 0.476 & 0.429 & 0.439 \\
            & 336 & \bestres{0.351} & \bestres{0.388} & 0.410 & 0.426 & 0.414 & 0.423 & 0.428 & 0.432 & \secondres{0.386} & \secondres{0.414} & 0.452 & 0.541 & 0.426 & 0.433 & 1.043 & 0.731 & 0.643 & 0.571 & 0.594 & 0.541 & 0.496 & 0.487 \\
            & 720 & \bestres{0.376} & \bestres{0.415} & 0.411 & 0.433 & \secondres{0.408} & \secondres{0.432} & 0.427 & 0.445 & 0.412 & 0.434 & 0.462 & 0.657 & 0.431 & 0.446 & 1.104 & 0.763 & 0.874 & 0.679 & 0.831 & 0.657 & 0.463 & 0.474 \\
            \rowcolor{tabhighlight}
            & {\textbf{Avg.}} & \bestres{0.333} & \bestres{0.376} & 0.373 & 0.400 & 0.367 & 0.396 & 0.383 & 0.406 & \secondres{0.364} & \secondres{0.395} & 0.414 & 0.496 & 0.387 & 0.407 & 0.942 & 0.683 & 0.611 & 0.549 & 0.558 & 0.515 & 0.436 & 0.449 \\
        \midrule
        \multirow{4}[0]{*}{ETTm1}
            & 96 & \bestres{0.274} & \bestres{0.325} & 0.325 & 0.361 & \secondres{0.318} & \secondres{0.356} & 0.334 & 0.368 & 0.320 & 0.357 & 0.338 & 0.375 & 0.329 & 0.367 & 0.404 & 0.426 & 0.364 & 0.387 & 0.345 & 0.372 & 0.379 & 0.419 \\
            & 192 & \bestres{0.317} & \bestres{0.353} & 0.375 & 0.389 & 0.362 & 0.383 & 0.377 & 0.391 & \secondres{0.361} & \secondres{0.381} & 0.374 & 0.387 & 0.367 & 0.385 & 0.450 & 0.451 & 0.398 & 0.404 & 0.380 & 0.389 & 0.426 & 0.441 \\
            & 336 & \bestres{0.355} & \bestres{0.378} & 0.405 & 0.412 & 0.395 & 0.407 & 0.426 & 0.420 & \secondres{0.390} & \secondres{0.404} & 0.410 & 0.411 & 0.399 & 0.410 & 0.532 & 0.515 & 0.428 & 0.425 & 0.413 & 0.413 & 0.445 & 0.459 \\
            & 720 & \bestres{0.429} & \bestres{0.418} & 0.466 & 0.447 & \secondres{0.452} & 0.441 & 0.491 & 0.459 & 0.454 & 0.441 & 0.478 & 0.450 & 0.454 & \secondres{0.439} & 0.666 & 0.589 & 0.487 & 0.461 & 0.474 & 0.453 & 0.543 & 0.490 \\
            \rowcolor{tabhighlight}
            & {\textbf{Avg.}} & \bestres{0.343} & \bestres{0.369} & 0.393 & 0.403 & 0.382 & 0.397 & 0.407 & 0.410 & \secondres{0.381} & \secondres{0.395} & 0.400 & 0.405 & 0.387 & 0.400 & 0.513 & 0.495 & 0.419 & 0.419 & 0.403 & 0.406 & 0.448 & 0.452 \\
        \midrule
        \multirow{4}[0]{*}{ETTm2}
            & 96 & \bestres{0.166} & \bestres{0.248} & 0.180 & 0.261 & \secondres{0.171} & \secondres{0.256} & 0.180 & 0.264 & 0.175 & 0.258 & 0.187 & 0.267 & 0.175 & 0.259 & 0.287 & 0.366 & 0.207 & 0.305 & 0.193 & 0.292 & 0.203 & 0.287 \\
            & 192 & \bestres{0.223} & \bestres{0.287} & 0.246 & 0.306 & \secondres{0.237} & \secondres{0.299} & 0.250 & 0.309 & \secondres{0.237} & \secondres{0.299} & 0.249 & 0.309 & 0.241 & 0.302 & 0.414 & 0.492 & 0.290 & 0.364 & 0.284 & 0.362 & 0.269 & 0.328 \\
            & 336 & \bestres{0.274} & \bestres{0.321} & 0.319 & 0.352 & \secondres{0.296} & \secondres{0.338} & 0.311 & 0.348 & 0.298 & 0.340 & 0.321 & 0.351 & 0.305 & 0.343 & 0.597 & 0.542 & 0.377 & 0.422 & 0.369 & 0.427 & 0.325 & 0.366 \\
            & 720 & \bestres{0.365} & \bestres{0.378} & 0.405 & 0.401 & 0.392 & \secondres{0.394} & 0.412 & 0.407 & \secondres{0.391} & 0.396 & 0.408 & 0.403 & 0.402 & 0.400 & 1.730 & 1.042 & 0.558 & 0.524 & 0.554 & 0.522 & 0.421 & 0.415 \\
            \rowcolor{tabhighlight}
            & {\textbf{Avg.}} & \bestres{0.257} & \bestres{0.308} & 0.287 & 0.330 & \secondres{0.274} & \secondres{0.322} & 0.288 & 0.332 & 0.275 & 0.323 & 0.291 & 0.332 & 0.281 & 0.326 & 0.757 & 0.610 & 0.358 & 0.403 & 0.350 & 0.400 & 0.304 & 0.349 \\
        \midrule
        \multirow{4}[0]{*}{Weather}
            & 96 & \bestres{0.146} & \bestres{0.188} & 0.166 & 0.208 & \secondres{0.157} & \secondres{0.205} & 0.174 & 0.214 & 0.163 & 0.209 & 0.172 & 0.220 & 0.177 & 0.218 & 0.158 & 0.230 & 0.202 & 0.261 & 0.196 & 0.255 & 0.217 & 0.296 \\
            & 192 & \bestres{0.190} & \bestres{0.231} & 0.217 & 0.253 & \secondres{0.204} & \secondres{0.247} & 0.221 & 0.254 & 0.208 & 0.250 & 0.219 & 0.261 & 0.225 & 0.259 & 0.206 & 0.277 & 0.242 & 0.298 & 0.237 & 0.296 & 0.276 & 0.336 \\
            & 336 & \bestres{0.242} & \bestres{0.271} & 0.282 & 0.300 & 0.261 & 0.290 & 0.278 & 0.296 & \secondres{0.251} & \secondres{0.287} & 0.280 & 0.306 & 0.278 & 0.297 & 0.272 & 0.335 & 0.287 & 0.335 & 0.283 & 0.335 & 0.339 & 0.380 \\
            & 720 & \bestres{0.314} & \bestres{0.324} & 0.356 & 0.351 & 0.340 & \secondres{0.341} & 0.358 & 0.349 & \secondres{0.339} & \secondres{0.341} & 0.365 & 0.359 & 0.354 & 0.348 & 0.398 & 0.418 & 0.351 & 0.386 & 0.345 & 0.381 & 0.403 & 0.428 \\
            \rowcolor{tabhighlight}
            & {\textbf{Avg.}} & \bestres{0.223} & \bestres{0.253} & 0.255 & 0.278 & 0.241 & \secondres{0.271} & 0.258 & 0.278 & \secondres{0.240} & \secondres{0.271} & 0.259 & 0.286 & 0.259 & 0.281 & 0.258 & 0.315 & 0.270 & 0.320 & 0.265 & 0.316 & 0.308 & 0.360 \\
        \midrule
        \multirow{4}[0]{*}{ECL}
            & 96 & \bestres{0.132} & \bestres{0.225} & 0.143 & \secondres{0.233} & \secondres{0.140} & 0.242 & 0.148 & 0.240 & 0.153 & 0.247 & 0.168 & 0.272 & 0.181 & 0.270 & 0.219 & 0.314 & 0.237 & 0.329 & 0.197 & 0.282 & 0.193 & 0.308 \\
            & 192 & \bestres{0.149} & \bestres{0.241} & 0.158 & \secondres{0.248} & \secondres{0.157} & 0.256 & 0.162 & 0.253 & 0.166 & 0.256 & 0.184 & 0.289 & 0.188 & 0.274 & 0.231 & 0.322 & 0.236 & 0.330 & 0.196 & 0.285 & 0.201 & 0.315 \\
            & 336 & \bestres{0.167} & \bestres{0.259} & 0.178 & \secondres{0.269} & \secondres{0.176} & 0.275 & 0.178 & \secondres{0.269} & 0.185 & 0.277 & 0.198 & 0.300 & 0.204 & 0.293 & 0.246 & 0.337 & 0.249 & 0.344 & 0.209 & 0.301 & 0.214 & 0.329 \\
            & 720 & \bestres{0.209} & \bestres{0.297} & 0.218 & \secondres{0.305} & \secondres{0.211} & 0.306 & 0.225 & 0.317 & 0.225 & 0.310 & 0.220 & 0.320 & 0.246 & 0.324 & 0.280 & 0.363 & 0.284 & 0.373 & 0.245 & 0.333 & 0.246 & 0.355 \\
            \rowcolor{tabhighlight}
            & {\textbf{Avg.}} & \bestres{0.164} & \bestres{0.255} & 0.174 & \secondres{0.264} & \secondres{0.171} & 0.270 & 0.178 & 0.270 & 0.182 & 0.272 & 0.192 & 0.295 & 0.205 & 0.290 & 0.244 & 0.334 & 0.251 & 0.344 & 0.212 & 0.300 & 0.214 & 0.327 \\
        \midrule
        \multirow{4}[0]{*}{Traffic} 
            & 96 & \bestres{0.367} & \bestres{0.248} & \secondres{0.376} & \secondres{0.251} & 0.428 & 0.271 & 0.395 & 0.268 & 0.462 & 0.285 & 0.593 & 0.321 & 0.462 & 0.295 & 0.522 & 0.290 & 0.805 & 0.493 & 0.650 & 0.396 & 0.587 & 0.366 \\
            & 192 & \bestres{0.384} & \bestres{0.256} & \secondres{0.398} & \secondres{0.261} & 0.448 & 0.282 & 0.417 & 0.276 & 0.473 & 0.296 & 0.617 & 0.336 & 0.466 & 0.296 & 0.530 & 0.293 & 0.756 & 0.474 & 0.598 & 0.370 & 0.604 & 0.373 \\
            & 336 & \bestres{0.398} & \bestres{0.266} & \secondres{0.415} & \secondres{0.269} & 0.473 & 0.289 & 0.433 & 0.283 & 0.498 & 0.296 & 0.629 & 0.336 & 0.482 & 0.304 & 0.558 & 0.305 & 0.762 & 0.477 & 0.605 & 0.373 & 0.621 & 0.383 \\
            & 720 & \bestres{0.439} & \secondres{0.295} & \secondres{0.447} & \bestres{0.287} & 0.516 & 0.307 & 0.467 & 0.302 & 0.506 & 0.313 & 0.640 & 0.350 & 0.514 & 0.322 & 0.589 & 0.328 & 0.719 & 0.449 & 0.645 & 0.394 & 0.626 & 0.382 \\
            \rowcolor{tabhighlight}
            & {\textbf{Avg.}} & \bestres{0.397} & \bestres{0.266} & \secondres{0.409} & \secondres{0.267} & 0.466 & 0.287 & 0.428 & 0.282 & 0.484 & 0.297 & 0.620 & 0.336 & 0.481 & 0.304 & 0.550 & 0.304 & 0.760 & 0.473 & 0.625 & 0.383 & 0.609 & 0.376 \\
        \midrule
        \multicolumn{2}{c|}{\scalebox{1.1}{\textbf{Average}}}
            & \bestres{0.300} & \bestres{0.320} & \secondres{0.334} & 0.341 & \secondres{0.334} & \secondres{0.340} & 0.342 & 0.346 & 0.339 & 0.342 & 0.376 & 0.371 & 0.353 & 0.352 & 0.542 & 0.466 & 0.458 & 0.431 & 0.410 & 0.396 & 0.394 & 0.396 \\
        \midrule
        \rowcolor{tabhighlight}\multicolumn{2}{c}{\textbf{{$1^{\text{st}}$ Count}}} 
            & \multicolumn{2}{c}{71} & \multicolumn{2}{c}{1} & \multicolumn{2}{c}{0} & \multicolumn{2}{c}{0} & \multicolumn{2}{c}{0} & \multicolumn{2}{c}{0} & \multicolumn{2}{c}{0} & \multicolumn{2}{c}{0} &  \multicolumn{2}{c}{0} & \multicolumn{2}{c}{0} & \multicolumn{2}{c}{0} \\
        \bottomrule
        \end{tabular}%
        }
        \vspace{-5pt}
    \end{table*}%

    We conduct extensive experiments to evaluate the effectiveness of \model layers in long-term multivariate forecasting. 
    Our evaluation framework compares \model against $15$ state-of-the-art baseline models from different architectures, including recent Transformer-based and non-Transformer approaches (see Appendix~\ref{apdx:baselines} for details). 
    We consider seven public benchmark datasets from diverse real-world domains (see Appendix~\ref{apdx:datasets} for a detailed overview of each dataset). 
    Forecasting accuracy is measured using mean squared error (MSE) and mean absolute error (MAE) on the held-out test sets as evaluation metrics (see definitions in Appendix~\ref{apdx:metrics}). 
    To ensure a fair comparison, we follow standard data processing and use a chronological train/validation/test splits to prevent information leakage~\cite{wu2023timesnet}. Each dataset is evaluated over four long-term forecasting horizons $H \in \{96, 192, 336, 720\}$.

\subsection{Multivariate Time Series Forecasting}
\label{subsec:forecasting}

    \Tabref{tab:ltsf_full} shows the long-term multivariate forecast performance on all benchmarks. In general, \model achieves state-of-the-art results, consistently outperforming recent competitive baselines. When comparing the average metric values on the horizons $\{96, 192, 336, 720\}$ (\textbf{Avg.} rows), \model yields substantial error reductions relative to the best-performing baselines on each dataset. For example, \model improves over TimeXer by $12.8\%$ in average MSE on ETTh1, over TimeMixer by $8.5\%$ on ETTh2 and $7.0\%$ on Weather, and over SOFTS by $1.8\%$ on Traffic.

    The advantage of \model is also significant at the longest horizon (i.e., predicting $720$ future time points), where forecast errors accumulate, and robustness becomes critical. In this setting, \model reduces MSE by $13.0\%$ compared to TimeXer on ETTh1 and by $7.8\%$ on ETTh2; in addition, it further improves over TimeMixer by $7.3\%$ on Weather. These gains indicate that incorporating segment-wise specialization enhances the model's ability to represent heterogeneous temporal patterns, especially for time-extended extrapolation. Additional comparisons with large time-series foundation models are provided in Appendix~\ref{apdx:results}.

\subsection{Ablation Study}
\label{subsec:ablation}

    \begin{table*}[!ht]
        \caption{Ablation study comparing the standard $\operatorname{MoE}$ and \model with different segment resolutions $\omega$. The best results are in \textbf{bold} and the second best are \underline{underlined}.}
        \label{tab:abl_smoe_vs_moe}
        \vspace{-2pt}
        \centering
        \begin{small}
        \resizebox{0.89\linewidth}{!}{
        \renewcommand{\multirowsetup}{\centering}
        \setlength{\tabcolsep}{6pt}
        \begin{tabular}{K{2.35cm}|cc|cc|cc|cc|cc|cc|cc}
        \toprule
            \multicolumn{1}{c}{\textbf{Dataset}} & \multicolumn{2}{c}{\textbf{ETTh1}} & \multicolumn{2}{c}{\textbf{ETTh2}} & \multicolumn{2}{c}{\textbf{ETTm1}} & \multicolumn{2}{c}{\textbf{ETTm2}} & \multicolumn{2}{c}{\textbf{Weather}} & \multicolumn{2}{c}{\textbf{ECL}} & \multicolumn{2}{c}{\textbf{Traffic}} \\
            \cmidrule(lr){2-3} \cmidrule(lr){4-5} \cmidrule(lr){6-7} \cmidrule(lr){8-9} \cmidrule(lr){10-11} \cmidrule(lr){12-13} \cmidrule(lr){14-15}
            \multicolumn{1}{c}{\textbf{Metrics (Avg.)}} & \textbf{MSE} & \textbf{MAE} & \textbf{MSE} & \textbf{MAE} & \textbf{MSE} & \textbf{MAE} & \textbf{MSE} & \textbf{MAE} & \textbf{MSE} & \textbf{MAE} & \textbf{MSE} & \textbf{MAE} & \textbf{MSE} & \textbf{MAE} \\
        \midrule
            $\operatorname{MoE}$ & 0.416 & 0.432 & 0.357 & 0.391 & 0.392 & 0.396 & 0.275 & 0.320 & 0.247 & 0.273 & 0.186 & \underline{0.276} & 0.442 & 0.303 \\
            \model $\omega=2$ & 0.402 & 0.419 & \underline{0.349} & \underline{0.384} & \textbf{0.360} & \underline{0.379} & 0.269 & \underline{0.315} & 0.240 & 0.268 & 0.187 & 0.281 & \underline{0.422} & 0.291 \\
            \model $\omega=3$ & \underline{0.396} & \underline{0.418} & 0.350 & 0.385 & 0.371 & 0.383 & \underline{0.267} & \underline{0.315} & \textbf{0.230} & \textbf{0.261} & 0.184 & 0.281 & \underline{0.422} & 0.289 \\
            \model $\omega=4$ & \underline{0.396} & \textbf{0.417} & \textbf{0.345} & \textbf{0.380} & 0.373 & 0.380 & \textbf{0.262} & 0.316 & \underline{0.236} & \underline{0.263} & \underline{0.179} & 0.277 & \textbf{0.415} & \underline{0.287} \\
            \model $\omega=5$ & \textbf{0.392} & \textbf{0.417} & 0.354 & \underline{0.384} & \underline{0.361} & \textbf{0.375} & \textbf{0.262} & \textbf{0.312} & \underline{0.236} & \underline{0.263} & \textbf{0.178} & \textbf{0.272} & \textbf{0.415} & \textbf{0.285} \\
        \bottomrule
        \end{tabular}
        }
        \end{small}
    \end{table*}

    \begin{table*}[!ht]
        \caption{Ablation study with multi-resolution \model. The best results are in \textbf{bold} and the second best are \underline{underlined}.}
        \label{tab:abl_multi_smoe}
        \vspace{-2pt}
        \centering
        \begin{small}
        \resizebox{0.89\linewidth}{!}{
        \renewcommand{\multirowsetup}{\centering}
        \setlength{\tabcolsep}{6pt}
        \begin{tabular}{K{2.35cm}|cc|cc|cc|cc|cc|cc|cc}
        \toprule
            \multicolumn{1}{c}{\textbf{Dataset}} & \multicolumn{2}{c}{\textbf{ETTh1}} & \multicolumn{2}{c}{\textbf{ETTh2}} & \multicolumn{2}{c}{\textbf{ETTm1}} & \multicolumn{2}{c}{\textbf{ETTm2}} & \multicolumn{2}{c}{\textbf{Weather}} & \multicolumn{2}{c}{\textbf{ECL}} & \multicolumn{2}{c}{\textbf{Traffic}} \\
            \cmidrule(lr){2-3} \cmidrule(lr){4-5} \cmidrule(lr){6-7} \cmidrule(lr){8-9} \cmidrule(lr){10-11} \cmidrule(lr){12-13} \cmidrule(lr){14-15}
            \multicolumn{1}{c}{\textbf{Metrics (Avg.)}} & \textbf{MSE} & \textbf{MAE} & \textbf{MSE} & \textbf{MAE} & \textbf{MSE} & \textbf{MAE} & \textbf{MSE} & \textbf{MAE} & \textbf{MSE} & \textbf{MAE} & \textbf{MSE} & \textbf{MAE} & \textbf{MSE} & \textbf{MAE} \\
        \midrule
            $\omega = [\,5,5,4,3\,]$ & 0.393 & 0.417 & 0.349 & 0.383 & 0.399 & 0.386 & 0.261 & 0.316 & \textbf{0.228} & \textbf{0.257} & \underline{0.178} & \textbf{0.270} & \underline{0.416} & 0.292 \\
            $\omega = [\,5,5,3,1\,]$ & 0.394 & 0.416 & 0.342 & 0.380 & 0.367 & 0.380 & \textbf{0.256} & 0.313 & \underline{0.230} & \underline{0.260} & 0.181 & 0.278 & \textbf{0.415} & \textbf{0.281} \\
            $\omega = [\,5,4,3,2\,]$ & 0.391 & 0.417 & 0.348 & 0.382 & 0.380 & 0.382 & 0.264 & 0.315 & 0.238 & 0.262 & \textbf{0.175} & \textbf{0.270} & 0.420 & \underline{0.288} \\
            $\omega = [\,5,4,3,1\,]$ & 0.400 & 0.423 & 0.345 & 0.380 & 0.364 & \underline{0.377} & \underline{0.257} & \textbf{0.309} & 0.235 & 0.263 & \underline{0.178} & \underline{0.271} & 0.427 & 0.293 \\
            $\omega = [\,4,5,5,4\,]$ & \textbf{0.385} & \textbf{0.413} & \underline{0.341} & \underline{0.377} & 0.372 & \underline{0.377} & \underline{0.257} & \underline{0.311} & 0.242 & 0.265 & 0.181 & 0.278 & 0.426 & 0.293 \\
            $\omega = [\,3,5,5,5\,]$ & 0.388 & 0.416 & 0.352 & 0.383 & \textbf{0.350} & \textbf{0.374} & 0.267 & 0.315 & 0.231 & \underline{0.260} & 0.182 & 0.276 & 0.419 & 0.293 \\
            $\omega = [\,1,3,5,5\,]$ & \underline{0.386} & \underline{0.415} & \textbf{0.336} & \textbf{0.376} & \underline{0.351} & 0.378 & 0.267 & 0.316 & 0.232 & \underline{0.260} & 0.187 & 0.281 & 0.437 & 0.300 \\
        \bottomrule
        \end{tabular}
        }
        \end{small}
        \vspace{-5pt}
    \end{table*}

    Previous work demonstrated that sparse architectures improve the accuracy-efficiency trade-off of dense Transformer forecasters in long-term multivariate settings~\cite{shi2024time,liu2024moirai}. Therefore, we focus our ablation study on the effects of \model routing and segmentation choices by comparing multiple configurations on the same benchmark protocol. To save computational time, we fix the patch length $P=8$, model dimension to \dmodel $=128$ (see Section~\ref{subsec:settings}), $N = 4$ experts, $K = 1$ $\operatorname{Top-K}$ expert, and limit training up to $20$ epochs for all ablation experiments. We report MSE and MAE metrics, averaged over horizons $H\in\{96,192,336,720\}$, with lower values indicating better forecasting performance.

    \vspace{1pt}\textbf{MoE vs. Uniform \model.} Table~\ref{tab:abl_smoe_vs_moe} shows that replacing token-wise with segment-wise routing yields consistent gains across all benchmarks. Most of the \model configurations ($\omega\in\{2,3,4,5\}$) improve over the standard $\operatorname{MoE}$, indicating that routing and processing contiguous contexts is a more influential inductive bias than point-wise expert assignment for long-term forecasting. These results support the hypothesis that segment-wise routing reduces sensitivity to noisy individual patches and enables experts to specialize in coherent local patterns that span multiple adjacent patches. The standard $\operatorname{MoE}$ achieves only a single second-best entry for ECL MAE; however, \model outperforms it with the best MSE and MAE at $\omega=5$ for ECL. In summary, this ablation study isolates the benefit of uniform segment-wise routing and shows that increasing segment resolution generally strengthens performance for long-term multivariate forecasting.
    
    \vspace{1pt}\textbf{Multi-Resolution \model.} We evaluate multi-resolution \model variants to assess whether layer-wise granularity yields additional gains beyond uniform segmentation.
    Table~\ref{tab:abl_multi_smoe} reports the best metric values observed on a comprehensive segment-resolution sweep.
    Overall, multi-resolution routing consistently outperforms the token-wise $\operatorname{MoE}$ baseline and also strengthens the best uniform \model variants (see Table~\ref{tab:abl_smoe_vs_moe}) across virtually all configurations, with clear margins when comparing the best entries per dataset. These results confirm that (\textit{i}) segment-wise routing remains beneficial when the model is allowed to use different granularities across depth, and (\textit{ii}) adding a temporal hierarchy to routing provides an additional element of improvement in forecasting.
    
    At the same time, no single segment schedule dominates across all datasets.
    The best-performing configurations vary by domain, consistent with the heterogeneity of multivariate time series (i.e., different datasets exhibit distinct mixtures of periodicity, trend dynamics, and noise levels).
    Empirically, we achieved improved results by combining coarse-dominant segments (most layers with $\omega\in\{3,4,5\}$) with a few lower-resolution layers ($\omega\in\{1,2\}$).
    These patterns support the intended role of multi-resolution segment processing: fine-grained layers capture local structures (e.g., abrupt changes and local irregularities), while larger segments aggregate more contextual information (e.g., stable cycles), allowing the model to allocate capacity across different temporal scales at different depths.

\section{Conclusion}
\label{sec:conclusion}

    The success of Mixture-of-Experts in natural language processing has long relied on the assumption that individual tokens can be routed independently by specialized expert networks. However, time series typically exhibit patterns that rarely depend on isolated time steps but unfold over contiguous segments. Our proposed \model demonstrates that aligning this structural prior with the granularity of expert routing is not just an implementation detail but a key architectural inductive bias for sequential data. 
    By equipping a standard encoder-only $\operatorname{MoE}$ Transformer forecaster with \model layers, we elevated it to consistent state-of-the-art performance in long-term multivariate forecasting. Moreover, enabling multi-resolution segment routing and processing across Transformer blocks further enhances the model's robustness to dynamic, multi-scale temporal patterns. Therefore, \model not only advances sequential data modeling, but also opens new avenues for more flexible and powerful architectures, paving the way for larger, more semantically aware models in time series forecasting and general sequential modeling.

\section*{Acknowledgments}
    
    This work was supported in part by the Paulo Pinheiro de Andrade Fellowship.
    The opinions, hypotheses, conclusions, or recommendations expressed in this material are the authors' responsibility and do not necessarily reflect the views of the funding agencies.

\section*{Impact Statement}

    This work advances long-term multivariate time-series forecasting by proposing segment-wise $\operatorname{MoE}$ layers, which significantly improve predictive accuracy and efficiency. Time-series forecasting is a critical task in multiple domains. In general, improvements in prediction accuracy and speed can theoretically support positive social outcomes, such as anticipating future trends and optimizing private or public planning. However, forecasting models also carry risks. Like other Artificial Intelligence systems, they can reproduce or amplify biases present in historical data. To mitigate these risks, we emphasize the need for rigorous validation, clear documentation of data provenance, and the participation of human experts in the deployment pipeline. We do not anticipate any malicious uses specific to this research beyond the usual concerns for automated forecasting systems, but we encourage practitioners to apply domain-specific safeguards and ethical guidelines before operational deployment.

\bibliography{ref}
\bibliographystyle{icml2026}

\newpage
\appendix
\onecolumn

\section{Technical Details}
\label{apdx:technical}

\subsection{Transformer Model Components}
\label{apdx:components}

    \textbf{Channel Independence and Input Patch Embedding.} We first apply instance normalization to mitigate non-stationarity issues \cite{liu2022non}. Then, we process each channel (variate) of the multivariate data as an independent univariate series, applying it in parallel to a shared input embedding and Transformer weights across all channels, an approach known as channel independence \cite{wang2024timexer}. \citet{nie2022time} demonstrated that channel independence enables $n$-dimensional data handling and contributes to convergence by reducing overfitting. Furthermore, each input time series is segmented into fixed-length, non-overlapping subseries or ``patches.'' In practice, the input embedding layer reshapes a look-back window of length $L$ into $M = \lceil L / P \rceil$ non-overlapping patches of length $P$ per channel \cite{wang2024timexer,liu2024moirai}. This patching mechanism reduces the sequence length, and thus the memory-complexity cost of attention by a factor of $P$. \citet{nie2022time} showed that patching retains local structure while allowing longer look-back windows.

    \begin{wrapfigure}{r}{0.35\textwidth}
        \centering
        \vspace{12pt}
        \includegraphics[width=0.24\textwidth]{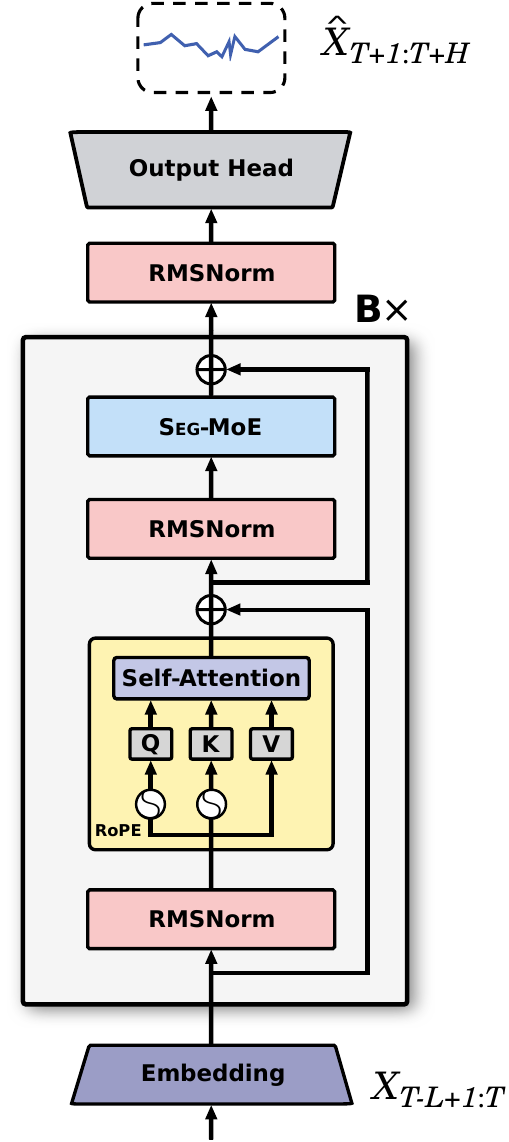}
        \caption{Encoder-only Transformer architecture used to experiment with \model layers in time series forecasting.}
        \label{fig:transformer}
    \end{wrapfigure}

    \vspace{0.1cm}\textbf{Transformer Encoder Blocks.} As the core time-series encoder backbone, we use the encoder-only Transformer architecture, composed of a stack of Transformer blocks~\cite{vaswani2017attention}. Each block applies two sequential sub-layers: (\textit{i}) an RMS normalization~\cite{zhang2019root} followed by a multi-head self-attention; and (\textit{ii}) an RMS normalization followed by a \model replacing the standard feed-forward network. Residual connections are set around each sub-layer to improve gradient flow. We use Pre-Norm instead of Post-Norm \cite{xiong2020layer} and replace standard $\operatorname{LayerNorm}$~\cite{ba2016layer} with $\operatorname{RMSNorm}$ for efficiency, as $\operatorname{RMSNorm}$ has been shown to achieve similar accuracy to $\operatorname{LayerNorm}$ while being more computationally efficient~\cite{zhang2019root,grattafiori2024llama}. Figure~\ref{fig:transformer} illustrates the Transformer architecture that we integrate with \model.

    \vspace{0.1cm}\textbf{Positional Encoding and Self-Attention.} The self-attention sub-layer uses Grouped-Query Attention (GQA) to balance computational efficiency and modeling capacity~\cite{ainslie2023gqa}. Instead of the standard multi-head attention, in which query, key, and value projections have the same number of heads~\cite{vaswani2017attention}, GQA clusters query heads into tunable groups that share a single key/value projection. We use a light grouping factor of 2, i.e., 2 query heads per key/value head, which halves the number of key/value matrices without significantly degrading accuracy compared to the standard multi-head attention~\cite{ainslie2023gqa}.
    Moreover, we apply Rotary Positional Embeddings (RoPE)~\cite{su2024roformer} to each query and key projections to encode temporal order. RoPE is a relative positional approach that allows the model to generalize to variable sequence lengths while capturing decaying dependencies with distance.
    Finally, we compute scaled-dot product attention using FlashAttention~\cite{dao2022flashattention}, which optimizes GPU memory reads and writes to reduce memory overhead and accelerate training.

    \vspace{0.1cm}\textbf{Feed-Forward and \model.} Standard Transformer blocks apply a two-layer feed-forward network ($\operatorname{FFN}$) after the attention sub-layer~\cite{vaswani2017attention}. We replace this dense $\operatorname{FFN}$-based configuration with our proposed segment-wise Mixture-of-Experts (\model) layer (see Section~\ref{subsec:segmoe}) to introduce sparsity and enhance model capacity in time-series forecasting tasks.

\subsection{\model Components}
\label{apdx:moe_components}
    
    \textbf{Routing Mechanism.} Our segment-wise Mixture-of-Experts (\model) uses a single global (shared) expert alongside multiple specialized, routed experts, a composition designed to capture time patterns at the segment level~\cite{dai2024deepseekmoe}. All expert networks are initialized with random values. During training optimization, the learnable routing mechanism learns to assign similar time segments to the same expert. Only those experts selected for a given segment receive gradient updates, which allows them to specialize in processing that type of time pattern~\cite{shi2024time,liu2024moirai}. This routing is performed independently at each \model layer throughout the model, resulting in a sparse expert structure, each specialized to handle different time characteristics.
    
    Specifically, for a given input segment $c$ of length $W$, the shared expert's contribution is modulated by a learnable $\operatorname{Sigmoid}$-based gating function $g_{N+1,c}$~\cite{shi2024time}. In parallel, the routing mechanism selects $K$ routed experts (from $N$ $\operatorname{FFN}$ expert networks) according to the $\operatorname{Top-K}$ routing weights $g_{i,c}$ (Equation~\ref{equ:expert_score}) computed from the $\operatorname{Softmax}$ probabilities $s_{i,c}$ over the learnable gating projection $\textbf{G}_i$ (Equation~\ref{equ:expert_gate}). Only the $K$ highest scores are retained, with all others set to zero~\cite{dai2024deepseekmoe,gshard}. The output of a \model layer is a weighted combination of the shared expert and the selected routed experts, ensuring sparsity. As \model extends the capacities of a standard $\operatorname{MoE}$, it also enables models to scale without a proportional increase in computation by activating only a subset of experts per time segment.

\subsection{Loss Functions}
\label{apdx:loss}

    \textbf{Prediction Loss:} We use the Huber loss. For a predicted time point $\hat{\mathbf{x}}_t$ and ground truth $\mathbf{x}_t$, the Huber loss is defined as:
    \vspace{5pt}
    \begin{equation}
        \mathcal{L}_{\text{pred}} ( \mathbf{x}_t, \hat{\mathbf{x}}_t ) =
        \begin{cases}
        0.5 ( \mathbf{x}_t - \hat{\mathbf{x}}_t )^{2}, \quad\quad\quad\quad\quad \text{if } \left| \mathbf{x}_t - \hat{\mathbf{x}}_t \right| \leq \delta,\\ \label{equ:huber}
        \delta \times ( \left| \mathbf{x}_t - \hat{\mathbf{x}}_t \right| - 0.5 \times \delta ), \text{ otherwise},
        \end{cases}
    \end{equation}
    
    with $\delta$ being a hyperparameter that balances the quadratic (MSE) and linear (L1) regimes. Recent work demonstrated that using the Huber loss for training time-series forecasting models yields better performance than using only MSE or L1 losses, due to the Huber loss's robustness to outliers, which are common in high-frequency data~\cite{shi2024time}.

    \vspace{0.1cm}\textbf{Expert Routing-Balance Loss.} Sparse $\operatorname{MoE}$ architectures, including our segment-wise $\operatorname{MoE}$, rely on learned routing mechanisms that are prone to load imbalance, in which the routing mechanism learns to assign most segments to a few experts or even a single expert. This phenomenon causes a ``routing collapse,'' in which a few experts are overused while the others are rarely selected, reducing the overall capacity and specialization advantages of the $\operatorname{MoE}$ approach~\cite{shi2024time,dai2024deepseekmoe,shazeer2017outrageously,fedus2022switch}. To mitigate this issue, we incorporate the auxiliary expert routing-balance loss proposed by \citet{fedus2022switch}. The expert routing-balance loss penalizes experts that receive disproportionately high gating scores, encouraging a more balanced distribution of input segments across all experts. In summary, the expert routing-balance loss (Equation~\ref{equ:aux_loss}) is computed as the fraction of time segments $f_i$ routed to expert $i$ multiplied by its routing probability $r_i$. By assigning a higher penalty to experts with larger $f_i r_i$ values, this simple formulation encourages uniform expert utilization and helps prevent any single expert from monopolizing the input data traffic during training.

\section{Experimental Details}
\label{apdx:details}

\subsection{Baseline Models}
\label{apdx:baselines}

    We compare against a set of $15$ state-of-the-art time series forecasting models from different architectures, including:

    \begin{itemize}
        \item \textbf{In-domain (full-shot) Transformer-based models:} TimeXer~\cite{wang2024timexer}, iTransformer~\cite{liu2023itransformer}, TimesNet~\cite{wu2023timesnet}, PatchTST~\cite{nie2022time}, Crossformer~\cite{zhang2022crossformer}, and FEDformer~\cite{zhou2022fedformer}.

        \item \textbf{In-domain (full-shot) MLP/Convolutional forecasters:} SOFTS~\cite{han2024softs}, TimeMixer~\cite{wangtimemixer}, TiDE~\cite{das2023long}, and DLinear~\cite{zeng2023transformers}.

        \item \textbf{Large pre-trained (foundation) models:} Timer-XL~\cite{liu2024timerxl}, Time-MoE~\cite{shi2024time}, Moirai~\cite{woo2024moirai}, MOMENT~\cite{goswami2024moment}, and Chronos~\cite{ansari2024chronos}.
    \end{itemize}

    In particular, PatchTST applies channel-independent patching to extend attention to long horizons, Crossformer explicitly models inter-variable dependencies, TimeMixer mixes multiple temporal scales, DLinear uses a series decomposition followed by independent linear layers, SOFTS is an efficient MLP model that aggregates all series into a global ``core'' representation and then redistributes it to capture cross-channel correlations. 
    Time series foundation models have demonstrated strong performance. In this context, Timer-XL is a unified causal decoder-only Transformer pre-trained on large corpora to enable zero-shot forecasting, Chronos uses a tokenized time-series approach trained on multiple datasets, and Time-MoE is a sparse Mixture-of-Experts Transformer family (up to 2.4B parameters) trained on over $300$ billion points to improve scalability. 
    We report the official results from~\cite{liu2023itransformer,wang2024timexer,han2024softs,liu2024timer}.

\subsection{Dataset Descriptions}
\label{apdx:datasets}

    We evaluate the performance of \model on seven real-world multivariate forecasting benchmarks. The Electricity Transformer Temperature (ETT) collection~\cite{zhou2021informer} includes four subsets (ETTh1/h2 and ETTm1/m2) that capture seven variables related to electric power load metrics, recorded over two years in China and sampled at hourly (ETTh1/h2) and 15-minute (ETTm1/m2) resolutions. The Jena Weather dataset~\cite{wu2021autoformer} includes $21$ meteorological features (e.g., humidity, pressure, and temperature) sampled every $10$ minutes during 2020. We also use the Electricity Consumption (ECL) dataset~\cite{wu2021autoformer}, which aggregates hourly electricity usage for $321$ customers (recorded at 15-minute intervals between 2012 and 2014), and the Traffic dataset~\cite{wu2021autoformer}, which reports hourly road occupancy rates from $862$ traffic sensors in the San Francisco Bay Area between 2015 and 2016. 

    These datasets encompass a variety of domains with different temporal resolutions and seasonality structures. All of them are publicly available and widely used in the time series modeling literature. 
    \Tabref{tab:dataset} summarizes the dimensionality, split sizes, and a ``forecastability'' score for each dataset.

    \begin{table}[!ht]
    \centering
    \caption{Main statistics of each benchmark dataset. Dim denotes the number of variables. Dataset size refers to the number of time points and is organized into (Train, Validation, Test) splits.}
    \label{tab:dataset}
    \resizebox{0.86\columnwidth}{!}{
        \begin{small}
        \renewcommand{\multirowsetup}{\centering}
        \setlength{\tabcolsep}{3.8pt}
        \begin{tabular}{c|c|c|c|c|c|c}
        \toprule
            Task & Dataset & Dim & Dataset Size & Frequency & Forecastability & Information \\
        \midrule
        \multirow{6}{*}[-1.8em]{\begin{tabular}[c]{@{}c@{}}Long-term\\Forecasting\end{tabular}}
            & ETTh1 & 7 & (8545, 2881, 2881) & 1 Hour & 0.38 & Power Load \\
        \cmidrule{2-7}
            & ETTh2 & 7 & (8545, 2881, 2881) & 1 Hour & 0.45 & Power Load \\
        \cmidrule{2-7}
            & ETTm1 & 7 & (34465, 11521, 11521) & 15 Min & 0.46 & Power Load \\
        \cmidrule{2-7}
            & ETTm2 & 7 & (34465, 11521, 11521) & 15 Min & 0.55 & Power Load \\
        \cmidrule{2-7}
            & Weather & 21 & (36792, 5271, 10540) & 10 Min & 0.75  & Climate \\
        \cmidrule{2-7}
            & ECL & 321 & (18317, 2633, 5261) & 1 Hour & 0.77 & Electricity \\
        \cmidrule{2-7}
            & Traffic & 862 & (12185, 1757, 3509) & 1 Hour & 0.68  & Road Occupancy \\
        \bottomrule
        \end{tabular}
        \end{small}
    }
    \end{table}

    Forecastability is a rough measure of predictability computed as one minus the normalized spectral entropy of a time series~\cite{goerg2013forecastable}. Higher values indicate better predictability.

\subsection{Evaluation Metrics}
\label{apdx:metrics}

    We evaluate the forecasting performance of all models using Mean Squared Error (MSE) and Mean Absolute Error (MAE) over the prediction horizon $H$. These metrics are defined as follows:
    \begin{align}
        \label{eq:metrics}
        \text{MSE} &= \frac{1}{H}\sum_{t=1}^H (\mathbf{x}_{t} - \widehat{\mathbf{x}}_{t})^2,
        &
        \text{MAE} &= \frac{1}{H}\sum_{t=1}^H|\mathbf{x}_{t} - \widehat{\mathbf{x}}_{t}|,
    \end{align}
    where $\mathbf{x}_{t} \in \mathbb{R}$ is the ground-truth value, and $\widehat{\mathbf{x}}_{t} \in \mathbb{R}$ is the corresponding model's prediction at time step $t$. 
    For multivariate series, we compute each metric per variable and report the average across all dimensions. All results are reported on the held-out test split.

\section{Hyperparameter Settings}
\label{apdx:settings}

    Based on the configurations and training protocol described in Section~\ref{subsec:settings}, we defined two encoder-only Transformer backbones in a range of compute and capacity to experiment with our \model: 
    \smodel, with embedding dimension \dmodel $=128$ and experts' hidden dimension \dff $=256$; 
    and \largemodel, with embedding dimension \dmodel $=256$ and experts' hidden dimension \dff $=512$.
    All configurations are designed to support efficient inference on CPU hardware, remaining substantially lighter than recent large-scale long-horizon forecasters and time-series foundation models~\cite{shi2024time,liu2024moirai,das2023decoder,liu2024timerxl}. The corresponding architectural hyperparameters are summarized in Table~\ref{tab:backbone_model}. 
    Other settings can be easily defined simply by tuning the hyperparameters of the backbone or \model layers. 
    The number of activated parameters and the total number of parameters for each model setting vary according to the segment resolution schedule $\omega$. Table~\ref{tab:model_size} presents model sizes for uniform-resolution \model layers. Multi-resolution models are upper-bounded by the highest-resolution layers, meaning that mixing segment resolutions also helps reduce the number of activated parameters.

    \begin{table}[!ht]
        \caption{Summary of configurations for each Transformer encoder used as backbone to experiment with \model.}
        \label{tab:backbone_model}%
        \centering
        \begin{small}
        \resizebox{0.66\columnwidth}{!}{
        \begin{tabular}{lccccccc}
            \toprule
                Backbone Size & Blocks & Q-Heads & KV-Heads & Experts & $K$ & \dmodel & \dff \\
            \midrule
                Small & 4 & 4 & 2 & 4 & 1 & 128 & 256 \\ 
                Base  & 6 & 8 & 4 & 8 & 1 & 256 & 512 \\
            \bottomrule
        \end{tabular}%
        }
        \end{small}
    \end{table}%

    \begin{table}[!ht]
        \caption{Number of activated parameters and the total number of parameters of \model models according to the segment resolution schedule $\omega$.}
        \label{tab:model_size}%
        \centering
        \begin{small}
        \resizebox{0.62\columnwidth}{!}{
        \begin{tabular}{lcccccc}
            \toprule
                & Backbone Type & Activated Params & Total Params \\
            \midrule
                \smodel $\omega=2$ & Small & 1.7 $\mathrm{M}$ & 2.4 $\mathrm{M}$ \\ 
                \smodel $\omega=3$ & Small & 3.0 $\mathrm{M}$ & 3.8 $\mathrm{M}$ \\ 
                \smodel $\omega=4$ & Small & 4.8 $\mathrm{M}$ & 5.6 $\mathrm{M}$ \\ 
                \smodel $\omega=5$ & Small & 7.2 $\mathrm{M}$ & 7.9 $\mathrm{M}$ \\ 
                \basemodel $\omega=2$ & Base & 9.7 $\mathrm{M}$ & 20.7 $\mathrm{M}$ \\ 
                \basemodel $\omega=3$ & Base & 17.5 $\mathrm{M}$ & 28.6 $\mathrm{M}$ \\ 
                \basemodel $\omega=4$ & Base & 28.5 $\mathrm{M}$ & 39.6 $\mathrm{M}$ \\ 
                \basemodel $\omega=5$ & Base & 42.7 $\mathrm{M}$ & 53.8 $\mathrm{M}$ \\ 
            \bottomrule
        \end{tabular}%
        }
        \end{small}
    \end{table}%

    For regularization, we apply DropPath~\cite{huang2016deep} to the outputs of the attention and \model sub-layers, with a stochastic decay schedule that increases the dropping probability from shallow to deep blocks up to a maximum of $0.3$. We also use the standard Dropout~\cite{srivastava2014dropout} with a probability of $0.2$ across the remaining Transformer components~\cite{vaswani2017attention}. Unless otherwise stated, all learnable parameters are initialized using the xavier$\_$uniform initialization~\cite{glorot2010understanding}. 
    The code is publicly available on GitHub.

\section{Additional Experimental Results}
\label{apdx:results}

\subsection{Additional Baselines}
\label{apdx:other_baselines}

    To complement the main results against full-shot forecasters, we report additional results against large pre-trained time-series foundation models in Table~\ref{tab:ltsf_additional}. Specifically, we include Timer-XL, Time-MoE, Moirai, MOMENT, and Chronos (see Section~\ref{apdx:baselines}).
    Foundation models leverage large-scale pre-training on heterogeneous time-series corpora and often have large parameter spaces. General-purpose representations learned from diverse pre-training data may enable extrapolation advantages, especially in small, coarse-grained benchmarks such as ETTh1 and ETTh2.
    Training from scratch with small datasets is challenging for Transformer-based forecasters, as even moderately sized models can overfit~\cite{nie2022time}. As shown in Table~\ref{tab:ltsf_additional}, Timer-XL, Time-MoE, and Moirai achieve competitive performance on ETTh1, ETTh2, and ETTm1, reinforcing the valuable robustness induced by broad pre-training for generalization.
    
    Time-MoE and Moirai score the best values $3$ times each (aggregating wins across different versions). Despite operating with lighter settings, \model-based forecasters outperform the foundation-model baselines across most benchmarks and horizons. In particular, on Weather, we observe a $12.9\%$ reduction in average MSE relative to both Time-MoE\textsubscript{ultra} and Timer-XL, highlighting that the proposed segment-wise routing and processing mechanism can deliver performance without relying on extreme scale or massive pre-training. These results emphasize that aligning routing granularity with temporal locality can enhance accuracy, even when competing with $2.4$-billion-parameter state-of-the-art models, such as Time-MoE\textsubscript{ultra}.

    We note that all \model results are obtained with no pre-training. Investigating large-scale pre-training and zero-shot forecasting for \model is a promising direction, but we leave it for future work. For readability and to avoid visual clutter, we consolidate the best-performing \model variants into a single column in Tables~\ref{tab:results_model_config} and~\ref{tab:ltsf_full}. The exact \model configuration and training settings used for each reported result are provided in Table~\ref{tab:results_model_config}.

    \begin{table}[!ht]
        \centering
        \caption{Additional experiments of long-term multivariate forecasting against foundation model baselines. A lower MSE or MAE indicates a better prediction. Results are obtained from~\cite{liu2024timer}. {\bestres{Bold red}}: is the best value, \secondres{underlined blue}: the second best. $1^{\text{st}}$ Count is the number of wins achieved by models across prediction lengths and datasets.}
        \label{tab:ltsf_additional}
        \resizebox{\columnwidth}{!}{
        \begin{threeparttable}
        \renewcommand{\tabcolsep}{3pt}
        \begin{tabular}{cr|cc|cc|cc|cc|cc|cc|cc|cc|cc|cc|cc}
        \toprule
            \multicolumn{2}{c}{\scalebox{1.1}{\textbf{Models}}} & 
            \multicolumn{2}{c}{\textbf{{\model}}} & 
            \multicolumn{2}{c}{\textbf{Timer-XL}\textsubscript{Base}} &
            \multicolumn{2}{c}{\textbf{Time-MoE}\textsubscript{Base}} &
            \multicolumn{2}{c}{\textbf{Time-MoE}\textsubscript{Large}} &
            \multicolumn{2}{c}{\textbf{Time-MoE}\textsubscript{Ultra}} &
            \multicolumn{2}{c}{\textbf{Moirai}\textsubscript{Small}} &
            \multicolumn{2}{c}{\textbf{Moirai}\textsubscript{Base}} &
            \multicolumn{2}{c}{\textbf{Moirai}\textsubscript{Large}} &
            \multicolumn{2}{c}{\textbf{MOMENT}} &
            \multicolumn{2}{c}{\textbf{Chronos}\textsubscript{Base}} &
            \multicolumn{2}{c}{\textbf{Chronos}\textsubscript{Large}} \\
    
            \cmidrule(lr){3-4} \cmidrule(lr){5-6}\cmidrule(lr){7-8} \cmidrule(lr){9-10}\cmidrule(lr){11-12}\cmidrule(lr){13-14}\cmidrule(lr){15-16}\cmidrule(lr){17-18}\cmidrule(lr){19-20} \cmidrule(lr){21-22} \cmidrule(lr){23-24}

            \multicolumn{2}{c}{\scalebox{1.1}{\textbf{Metrics}}} & \textbf{MSE} & \textbf{MAE} & \textbf{MSE} & \textbf{MAE} & \textbf{MSE} & \textbf{MAE} & \textbf{MSE} & \textbf{MAE} & \textbf{MSE} & \textbf{MAE} & \textbf{MSE} & \textbf{MAE} & \textbf{MSE} & \textbf{MAE} & \textbf{MSE} & \textbf{MAE} & \textbf{MSE} & \textbf{MAE} & \textbf{MSE} & \textbf{MAE} & \textbf{MSE} & \textbf{MAE} \\
        \midrule
        \multirow{4}[1]{*}{ETTh1}
            & 96 & \bestres{0.343} & \secondres{0.381} & 0.369 & 0.391 & 0.357 & \secondres{0.381} & 0.350 & 0.382 & \secondres{0.349} & \bestres{0.379} & 0.401 & 0.402 & 0.376 & 0.392 & 0.381 & 0.388 & 0.688 & 0.557 & 0.440 & 0.393 & 0.441 & 0.390 \\
            & 192 & \bestres{0.378} & \secondres{0.405} & 0.405 & 0.413 & \secondres{0.384} & \bestres{0.404} & 0.388 & 0.412 & 0.395 & 0.413 & 0.435 & 0.421 & 0.412 & 0.413 & 0.434 & 0.415 & 0.688 & 0.560 & 0.492 & 0.426 & 0.502 & 0.524 \\
            & 336 & \bestres{0.394} & \bestres{0.419} & 0.418 & \secondres{0.423} & \secondres{0.411} & 0.434 & \secondres{0.411} & 0.430 & 0.447 & 0.453 & 0.438 & 0.434 & 0.433 & 0.428 & 0.485 & 0.445 & 0.675 & 0.563 & 0.550 & 0.462 & 0.576 & 0.467 \\
            & 720 & \bestres{0.408} & \bestres{0.441} & \secondres{0.423} & \bestres{0.441} & 0.449 & 0.477 & 0.427 & 0.455 & 0.457 & 0.462 & 0.439 & 0.454 & 0.447 & \secondres{0.444} & 0.611 & 0.510 & 0.683 & 0.585 & 0.882 & 0.591 & 0.835 & 0.583 \\
            \rowcolor{tabhighlight}
            & {\textbf{Avg.}} & \bestres{0.381} & \bestres{0.412} & 0.404 & \secondres{0.417} & 0.400 & 0.424 & \secondres{0.394} & 0.419 & 0.412 & 0.426 & 0.428 & 0.427 & 0.417 & 0.419 & 0.480 & 0.439 & 0.683 & 0.566 & 0.591 & 0.468 & 0.588 & 0.466 \\
        \midrule
        \multirow{4}[0]{*}{ETTh2}
            & 96 & \bestres{0.272} & \secondres{0.331} & \secondres{0.283} & 0.342 & 0.305 & 0.359 & 0.302 & 0.354 & 0.292 & 0.352 & 0.297 & 0.336 & 0.294 & \bestres{0.330} & 0.296 & \bestres{0.330} & 0.342 & 0.396 & 0.308 & 0.343 & 0.320 & 0.345 \\
            & 192 & \bestres{0.334} & \bestres{0.370} & \secondres{0.340} & 0.379 & 0.351 & 0.386 & 0.364 & 0.385 & 0.347 & 0.379 & 0.368 & 0.381 & 0.365 & 0.375 & 0.361 & \secondres{0.371} & 0.354 & 0.402 & 0.384 & 0.392 & 0.406 & 0.399 \\
            & 336 & \bestres{0.351} & \bestres{0.388} & 0.366 & 0.400 & 0.391 & 0.418 & 0.417 & 0.425 & 0.406 & 0.419 & 0.370 & 0.393 & 0.376 & \secondres{0.390} & 0.390 & \secondres{0.390} & \secondres{0.356} & 0.407 & 0.429 & 0.430 & 0.492 & 0.453 \\
            & 720 & \bestres{0.376} & \bestres{0.415} & 0.397 & 0.431 & 0.419 & 0.454 & 0.537 & 0.496 & 0.439 & 0.447 & 0.411 & 0.426 & 0.416 & 0.433 & 0.423 & \secondres{0.418} & \secondres{0.395} & 0.434 & 0.501 & 0.477 & 0.603 & 0.511 \\
            \rowcolor{tabhighlight}
            & {\textbf{Avg.}} & \bestres{0.333} & \bestres{0.376} & \secondres{0.347} & 0.388 & 0.366 & 0.404 & 0.405 & 0.415 & 0.371 & 0.399 & 0.361 & 0.384 & 0.362 & 0.382 & 0.367 & \secondres{0.377} & 0.361 & 0.409 & 0.405 & 0.410 & 0.455 & 0.427 \\
        \midrule
        \multirow{4}[0]{*}{ETTm1}
            & 96 & \bestres{0.274} & \bestres{0.325} & 0.317 & 0.356 & 0.338 & 0.368 & 0.309 & 0.357 & \secondres{0.281} & \secondres{0.341} & 0.418 & 0.392 & 0.363 & 0.356 & 0.380 & 0.361 & 0.654 & 0.527 & 0.454 & 0.408 & 0.457 & 0.403 \\
            & 192 & \secondres{0.317} & \bestres{0.353} & 0.358 & 0.381 & 0.353 & 0.388 & 0.346 & 0.381 & \bestres{0.305} & \secondres{0.358} & 0.431 & 0.405 & 0.388 & 0.375 & 0.412 & 0.383 & 0.662 & 0.532 & 0.567 & 0.477 & 0.530 & 0.450 \\
            & 336 & \bestres{0.355} & \bestres{0.378} & 0.386 & 0.401 & 0.381 & 0.413 & 0.373 & 0.408 & \secondres{0.369} & 0.395 & 0.433 & 0.412 & 0.416 & \secondres{0.392} & 0.436 & 0.400 & 0.672 & 0.537 & 0.662 & 0.525 & 0.577 & 0.481 \\
            & 720 & \bestres{0.429} & \bestres{0.418} & \secondres{0.430} & 0.431 & 0.504 & 0.493 & 0.475 & 0.477 & 0.469 & 0.472 & 0.462 & 0.432 & 0.460 & \bestres{0.418} & 0.462 & \secondres{0.420} & 0.692 & 0.551 & 0.900 & 0.591 & 0.660 & 0.526 \\
            \rowcolor{tabhighlight}
            & {\textbf{Avg.}} & \bestres{0.343} & \bestres{0.369} & 0.373 & 0.392 & 0.394 & 0.415 & 0.376 & 0.405 & \secondres{0.356} & 0.391 & 0.436 & 0.410 & 0.406 & \secondres{0.385} & 0.422 & 0.391 & 0.670 & 0.536 & 0.645 & 0.500 & 0.555 & 0.465 \\
        \midrule
        \multirow{4}[0]{*}{ETTm2}
            & 96 & \bestres{0.166} & \bestres{0.248} & \secondres{0.189} & 0.277 & 0.201 & 0.291 & 0.197 & 0.286 & 0.198 & 0.288 & 0.214 & 0.288 & 0.205 & 0.273 & 0.211 & 0.274 & 0.260 & 0.335 & 0.199 & 0.274 & 0.197 & \secondres{0.271} \\
            & 192 & \bestres{0.223} & \bestres{0.287} & 0.241 & 0.315 & 0.258 & 0.334 & 0.250 & 0.322 & \secondres{0.235} & \secondres{0.312} & 0.284 & 0.332 & 0.275 & 0.316 & 0.281 & 0.318 & 0.289 & 0.350 & 0.261 & 0.322 & 0.254 & 0.314 \\
            & 336 & \bestres{0.274} & \bestres{0.321} & \secondres{0.286} & \secondres{0.348} & 0.324 & 0.373 & 0.337 & 0.375 & 0.293 & \secondres{0.348} & 0.331 & 0.362 & 0.329 & 0.350 & 0.341 & 0.355 & 0.324 & 0.369 & 0.326 & 0.366 & 0.313 & 0.353 \\
            & 720 & \bestres{0.365} & \bestres{0.378} & \secondres{0.375} & \secondres{0.402} & 0.488 & 0.464 & 0.480 & 0.461 & 0.427 & 0.428 & 0.402 & 0.408 & 0.437 & 0.411 & 0.485 & 0.428 & 0.394 & 0.409 & 0.455 & 0.439 & 0.416 & 0.415 \\
            \rowcolor{tabhighlight}
            & {\textbf{Avg.}} & \bestres{0.257} & \bestres{0.308} & \secondres{0.273} & \secondres{0.336} & 0.317 & 0.365 & 0.316 & 0.361 & 0.288 & 0.344 & 0.307 & 0.347 & 0.311 & 0.337 & 0.329 & 0.343 & 0.316 & 0.365 & 0.310 & 0.350 & 0.295 & 0.338 \\
        \midrule
        \multirow{4}[0]{*}{Weather}
            & 96 & \bestres{0.146} & \bestres{0.188} & 0.171 & 0.225 & 0.160 & 0.214 & 0.159 & 0.213 & \secondres{0.157} & \secondres{0.211} & 0.198 & 0.222 & 0.220 & 0.217 & 0.199 & \secondres{0.211} & 0.243 & 0.255 & 0.203 & 0.238 & 0.194 & 0.235 \\
            & 192 & \bestres{0.190} & \bestres{0.231} & 0.221 & 0.271 & 0.210 & 0.260 & 0.215 & 0.266 & \secondres{0.208} & 0.256 & 0.247 & 0.265 & 0.271 & 0.259 & 0.246 & \secondres{0.251} & 0.278 & 0.329 & 0.256 & 0.290 & 0.249 & 0.285 \\
            & 336 & \bestres{0.242} & \bestres{0.271} & 0.274 & 0.311 & 0.274 & 0.309 & 0.291 & 0.322 & \secondres{0.255} & \secondres{0.290} & 0.283 & 0.303 & 0.286 & 0.297 & 0.274 & 0.291 & 0.306 & 0.346 & 0.314 & 0.336 & 0.302 & 0.327 \\
            & 720 & \bestres{0.315} & \bestres{0.324} & 0.356 & 0.370 & 0.418 & 0.405 & 0.415 & 0.400 & 0.405 & 0.397 & 0.373 & 0.354 & 0.373 & 0.354 & \secondres{0.337} & \secondres{0.340} & 0.350 & 0.374 & 0.397 & 0.396 & 0.372 & 0.378 \\
            \rowcolor{tabhighlight}
            & {\textbf{Avg.}} & \bestres{0.223} & \bestres{0.253} & \secondres{0.256} & 0.294 & 0.265 & 0.297 & 0.270 & 0.300 & \secondres{0.256} & 0.288 & 0.275 & 0.286 & 0.287 & 0.281 & 0.264 & \secondres{0.273} & 0.294 & 0.326 & 0.292 & 0.315 & 0.279 & 0.306 \\
        \midrule
        \multirow{4}[0]{*}{ECL}
            & 96 & \bestres{0.132} & \bestres{0.225} & \secondres{0.141} & 0.237 & -- & -- & -- & -- & -- & -- & 0.189 & 0.280 & 0.160 & 0.250 & 0.153 & 0.241 & 0.745 & 0.680 & 0.154 & 0.231 & 0.152 & \secondres{0.229} \\
            & 192 & \bestres{0.149} & \bestres{0.241} & \secondres{0.159} & 0.254 & -- & -- & -- & -- & -- & -- & 0.205 & 0.292 & 0.175 & 0.263 & 0.169 & 0.255 & 0.755 & 0.683 & 0.179 & 0.254 & 0.172 & \secondres{0.250} \\
            & 336 & \bestres{0.167} & \bestres{0.259} & \secondres{0.177} & \secondres{0.272} & -- & -- & -- & -- & -- & -- & 0.221 & 0.307 & 0.187 & 0.277 & 0.187 & 0.273 & 0.766 & 0.687 & 0.214 & 0.284 & 0.203 & 0.276 \\
            & 720 & \bestres{0.209} & \bestres{0.297} & \secondres{0.219} & \secondres{0.308} & -- & -- & -- & -- & -- & -- & 0.258 & 0.335 & 0.228 & 0.309 & 0.237 & 0.313 & 0.794 & 0.696 & 0.311 & 0.346 & 0.289 & 0.337 \\
            \rowcolor{tabhighlight}
            & {\textbf{Avg.}} & \bestres{0.164} & \bestres{0.255} & \secondres{0.174} & 0.278 & -- & -- & -- & -- & -- & -- & 0.218 & 0.303 & 0.187 & 0.274 & 0.186 & \secondres{0.270} & 0.765 & 0.686 & 0.214 & 0.278 & 0.204 & 0.273 \\
        \midrule
        \multicolumn{2}{c|}{\scalebox{1.1}{\textbf{Average}}}
            & \bestres{0.283} & \bestres{0.329} & \secondres{0.305} & 0.351 & 0.348 & 0.381 & 0.352 & 0.380 & 0.337 & 0.370 & 0.338 & 0.360 & 0.328 & \secondres{0.346} & 0.341 & 0.349 & 0.515 & 0.481 & 0.410 & 0.387 & 0.396 & 0.379 \\
        \midrule
        \rowcolor{tabhighlight}\multicolumn{2}{c}{\textbf{{$1^{\text{st}}$ Count}}} 
            & \multicolumn{2}{c}{58} & \multicolumn{2}{c}{1} & \multicolumn{2}{c}{1} & \multicolumn{2}{c}{0} & \multicolumn{2}{c}{2} & \multicolumn{2}{c}{0} & \multicolumn{2}{c}{2} & \multicolumn{2}{c}{1} & \multicolumn{2}{c}{0} & \multicolumn{2}{c}{0} & \multicolumn{2}{c}{0} \\
        \bottomrule
        \end{tabular}%
        \begin{tablenotes}
            \item[] $\ast$ Dataset used for pre-training is not evaluated on corresponding models; dashes denote results (--).
            \item[] $\ast$ Traffic from PEMS~\cite{liu2022scinet} is typically used for pre-training large time-series models and is therefore not evaluated here.
        \end{tablenotes}
        \end{threeparttable}
        }
    \end{table}%

    \begin{table}[!ht]
        \caption{Experiment configuration of \model according to the main and additional results reported in Tables \ref{tab:ltsf_full} and \ref{tab:ltsf_additional}. LR denotes learning rate.}
        \label{tab:results_model_config}
        \centering
        \begin{small}
        \resizebox{0.82\columnwidth}{!}{
        \renewcommand{\multirowsetup}{\centering}
        \setlength{\tabcolsep}{6pt}
        \begin{tabular}{ccccccccc}
        \toprule
            \multirow{2}[0]{*}{Dataset} & \multirow{2}[0]{*}{Model version} & \multirow{2}[0]{*}{Resolution} & \multicolumn{6}{c}{Training Process} \\
        \cmidrule(lr){4-9}
            & & $\omega$ & $P$ & $H_o$ & LR & Min LR & Batch Size & Epochs\\
        \midrule
            ETTh1 & \smodel & $[\, 4,5,5,4 \,]$ & 8 & 32 & $3.2\mathrm{e}\text{-}{4}$ & $1.2\mathrm{e}\text{-}{4}$ & 256 & 20 \\
            ETTh2 & \smodel & $[\, 3,5,5,5 \,]$ & 8 & 32 & $3.2\mathrm{e}\text{-}{4}$ & $1.2\mathrm{e}\text{-}{4}$ & 256 & 20 \\
            ETTm1 & \smodel & $[\, 3,5,5,5 \,]$ & 8 & 32 & $3.2\mathrm{e}\text{-}{4}$ & $1.2\mathrm{e}\text{-}{5}$ & 128 & 20 \\
            ETTm2 & \smodel & $[\, 3,5,5,5 \,]$ & 16 & 24 & $3.2\mathrm{e}\text{-}{4}$ & $1.2\mathrm{e}\text{-}{5}$ & 256 & 20 \\
            Weather & \smodel & $[\, 3,5,5,5 \,]$ & 8 & 32 & $3.2\mathrm{e}\text{-}{4}$ & $1.2\mathrm{e}\text{-}{5}$ & 256 & 20 \\
            ECL & \basemodel & $[\, 5,5,4,4,3,3 \,]$ & 8 & 32 & $2.2\mathrm{e}\text{-}{5}$ & $3.2\mathrm{e}\text{-}{6}$ & 16 & 10 \\
            Traffic & \basemodel & $[\, 5,5,4,3,2,2 \,]$ & 8 & 32 & $3.2\mathrm{e}\text{-}{5}$ & $1.2\mathrm{e}\text{-}{6}$ & 8 & 10 \\
        \bottomrule
        \end{tabular}
        }
        \end{small}
    \end{table}

\subsection{Patch Length Influence}
\label{apdx:patch}

    Forecast performance and efficiency are often sensitive to the patch length $P$, because patching simultaneously controls the effective sequence length processed by the Transformer and the temporal granularity at which patterns are embedded. 
    A small $P$ produces more patch embeddings per input window, increasing attention cost and GPU memory demands, but preserves local structures that can be critical for modeling short-term transitions.
    In contrast, a large $P$ reduces the number of patches and improves memory demands. However, larger patch sizes can compress high-frequency patterns and degrade accuracy, particularly in datasets with short fluctuations.
    
    Previous studies on long-term forecasting have reported this trade-off and typically find that patch sizes in $[8, 24]$ provide an optimal balance between scalability and fidelity to temporal structures~\cite{zhang2022crossformer,nie2022time,wang2024timexer,liu2024moirai}.
    In our evaluations, we experiment with $P \in \{8, 16\}$. We observe that $P=8$ produces the majority of our best-performing configurations, suggesting that retaining a relatively fine temporal granularity remains beneficial. Although patch length remains an important hyperparameter, segment-wise routing can make the architecture more robust to $P$. By routing and transforming contiguous segments of patches, \model introduces an additional locality mechanism, enabling layers to capture longer interactions and enhance expert specialization.

\subsection{Memory Footprint}
\label{apdx:memory}

\begin{figure}[!htb]
        \centering
        \includegraphics[width=0.4\columnwidth]{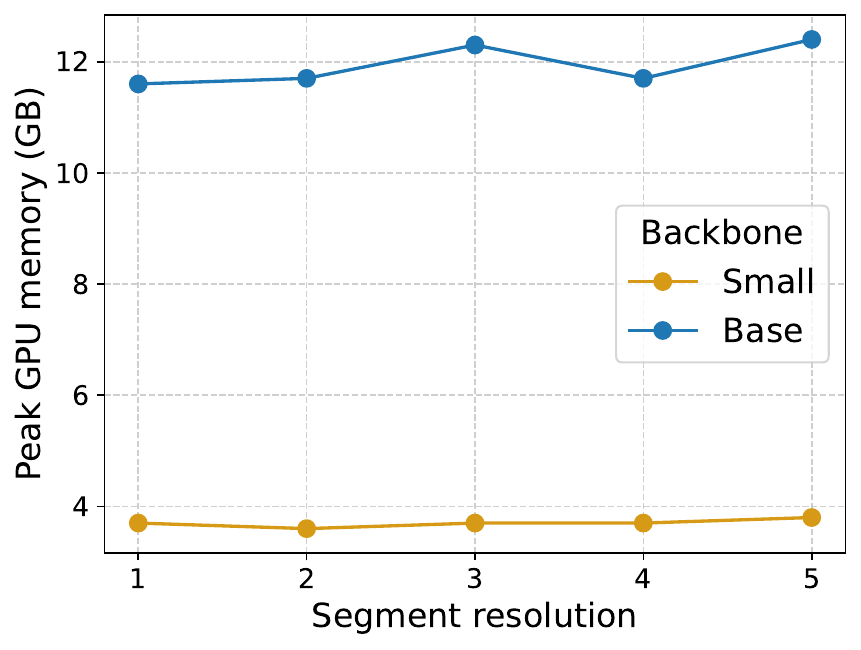}
        \caption{Memory footprint during training sweeping the segment resolution, where $\omega=1$ is equivalent to a standard $\operatorname{MoE}$ layer, and $\omega>1$ corresponds to our \model.}
        \label{fig:memory}
    \end{figure}

    We evaluate the memory footprint during training of \model relative to a standard token-wise $\operatorname{MoE}$ ($\omega=1$) under matched training conditions, considering the Transformer backbones defined in Section~\ref{apdx:settings}. For each backbone, we uniformly sweep the segment resolution $\omega \in \{1,2,3,4,5\}$, where $\omega=1$ corresponds to the standard $\operatorname{MoE}$, and $\omega>1$ to our \model. All experiments use a fixed patch length $P=8$ and a batch size of $128$, and we record the peak GPU memory allocated during training on the ETT benchmark datasets. Figure~\ref{fig:memory} illustrates memory demand as a function of segment size.

    Overall, the results indicate that segment-wise routing does not significantly increase the training memory footprint for \dmodel$=128$ (small) models and leads to a modest increase for larger model embeddings, such as \dmodel$=256$ (base). For the small backbone, peak memory remains essentially constant across segment resolutions (3.6--3.8 GB), with only a slight increase as $\omega$ increases (e.g., from 3.7 GB with $\omega=1$ and 3.8 GB with $\omega=5$). For the base backbone, the training memory ranges from 11.6 GB with $\omega=1$ to 12.4 GB with $\omega=5$, an increase of only $0.8$ GB.
    
    The effective batch processed by the model scales with the number of variables due to channel independence, and thus peak memory varies with dataset dimensionality and the chosen batch size. Nevertheless, in a controlled setting, \model achieves performance gains with comparable training memory requirements to token-wise $\operatorname{MoE}$, and the overhead of segment-wise routing is marginal for smaller backbones.

\section{Forecast Showcases}
\label{apdx:showcases}

\begin{figure}[!ht]
        \centering
        \begin{subfigure}{0.5\textwidth}
            \centering
            \includegraphics[width=0.8\linewidth]{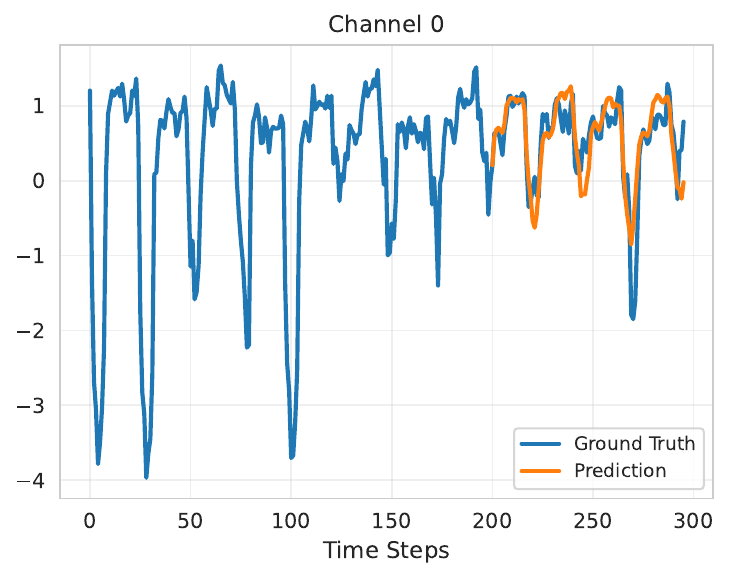}
            \caption{\model}
            \label{fig:show_segmoe_etth1}
        \end{subfigure}%
        \begin{subfigure}{0.5\textwidth}
            \centering
            \includegraphics[width=0.8\linewidth]{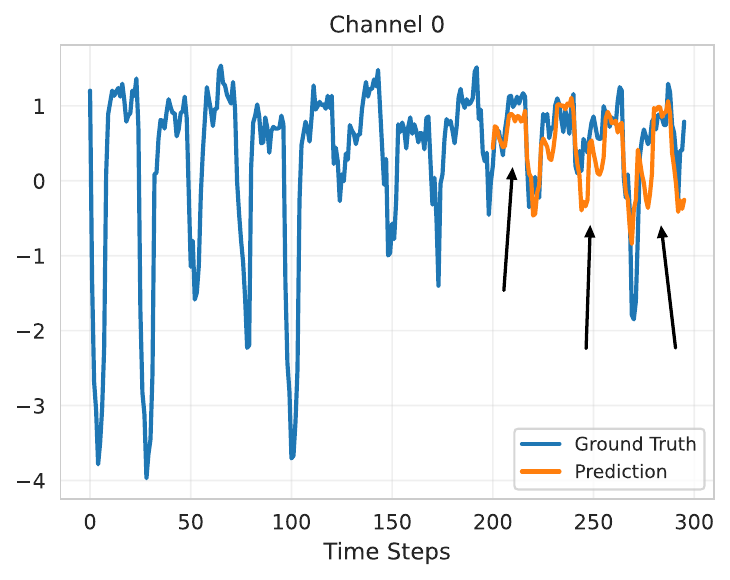}
            \caption{Standard MoE}
            \label{fig:show_moe_etth1}
        \end{subfigure}
        \caption{Forecast showcases from \textbf{ETTh1} with a forecast horizon of 96. Blue curves are the ground truths, and orange curves are the model predictions. The curves before the model predictions are the input data.}
        \label{fig:show_etth1}
    \end{figure}

\begin{figure}[!ht]
        \centering
        \begin{subfigure}{0.5\textwidth}
            \centering
            \includegraphics[width=0.8\textwidth]{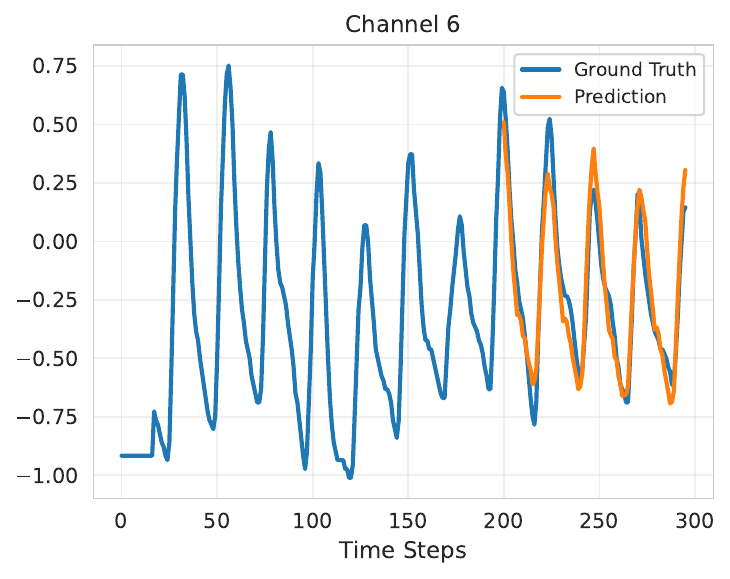}
            \caption{\model}
            \label{fig:show_segmoe_etth2}
        \end{subfigure}%
        \begin{subfigure}{0.5\textwidth}
            \centering
            \includegraphics[width=0.82\textwidth]{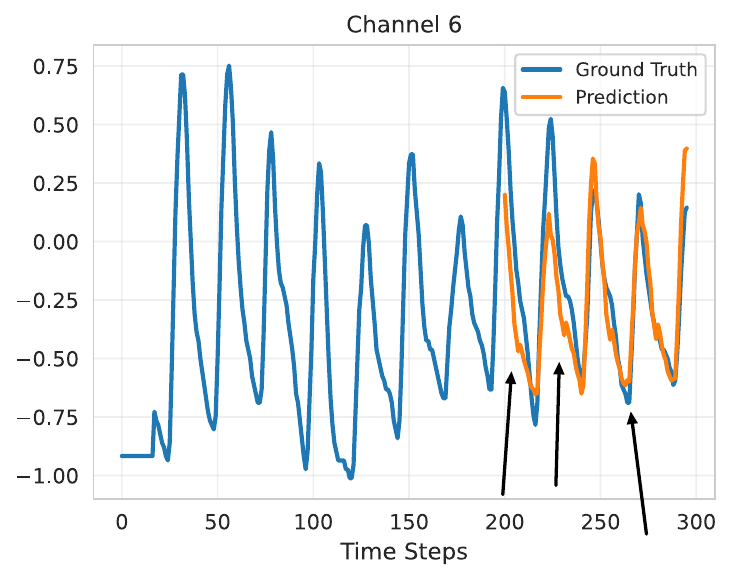}
            \caption{Standard MoE}
            \label{fig:show_moe_etth2}
        \end{subfigure}
        \caption{Forecast showcases from \textbf{ETTh2} with a forecast horizon of 96. Blue curves are the ground truths, and orange curves are the model predictions. The curves before the model predictions are the input data.}
        \label{fig:show_etth2}
    \end{figure}

\begin{figure}[!ht]
        \centering
        \begin{subfigure}{0.5\textwidth}
            \centering
            \includegraphics[width=0.82\textwidth]{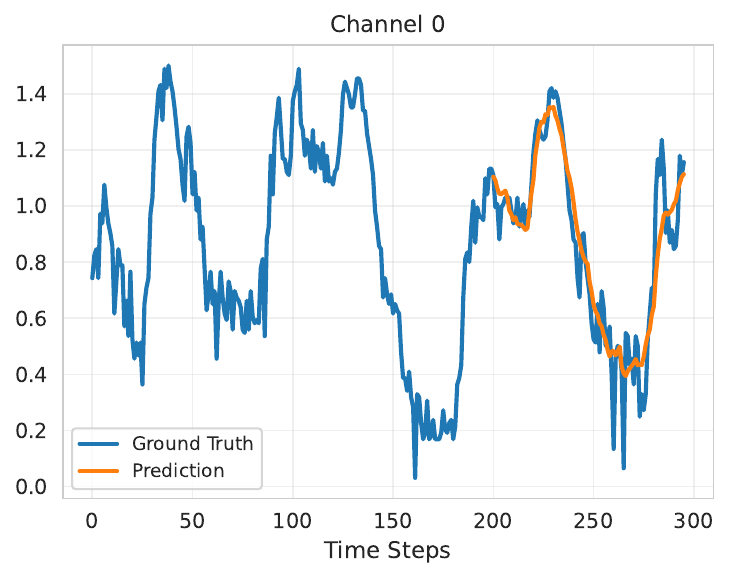}
            \caption{\model}
            \label{fig:show_segmoe_ettm1}
        \end{subfigure}%
        \begin{subfigure}{0.5\textwidth}
            \centering
            \includegraphics[width=0.82\textwidth]{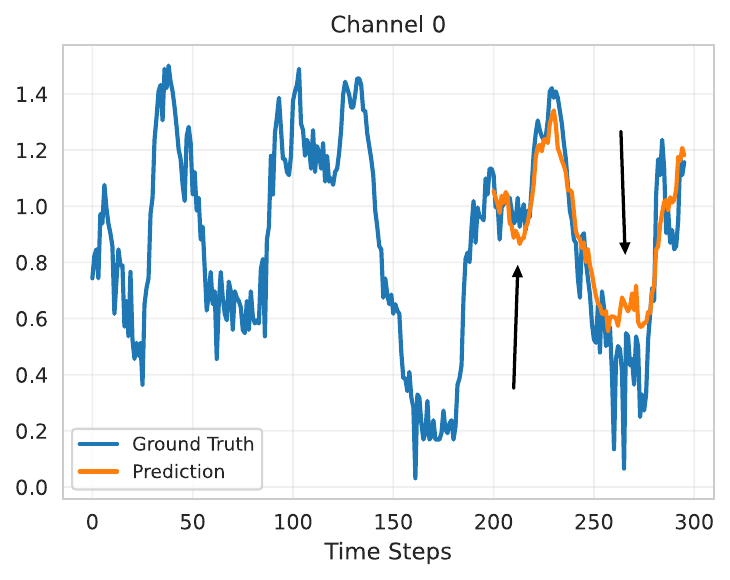}
            \caption{Standard MoE}
            \label{fig:show_moe_ettm1}
        \end{subfigure}
        \caption{Forecast showcases from \textbf{ETTm1} with a forecast horizon of 96. Blue curves are the ground truths, and orange curves are the model predictions. The curves before the model predictions are the input data.}
        \label{fig:show_ettm1}
    \end{figure}

\begin{figure}[!ht]
        \centering
        \begin{subfigure}{0.5\textwidth}
            \centering
            \includegraphics[width=0.82\textwidth]{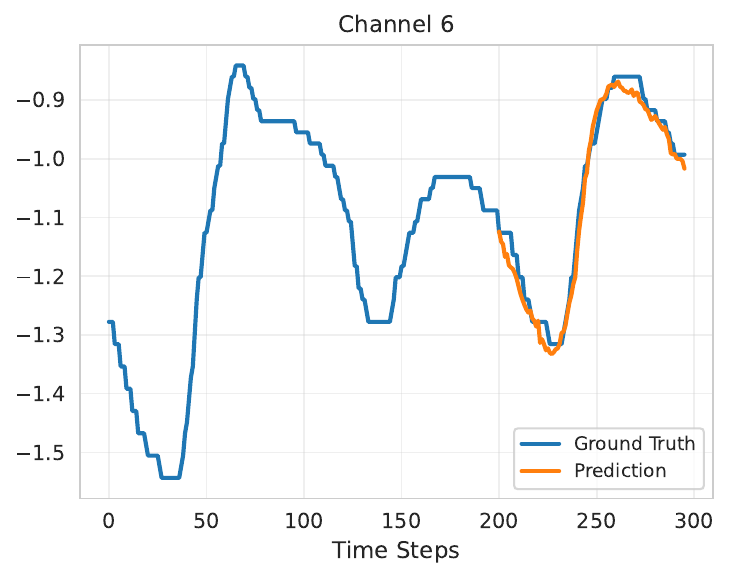}
            \caption{\model}
            \label{fig:show_segmoe_ettm2}
        \end{subfigure}%
        \begin{subfigure}{0.5\textwidth}
            \centering
            \includegraphics[width=0.82\textwidth]{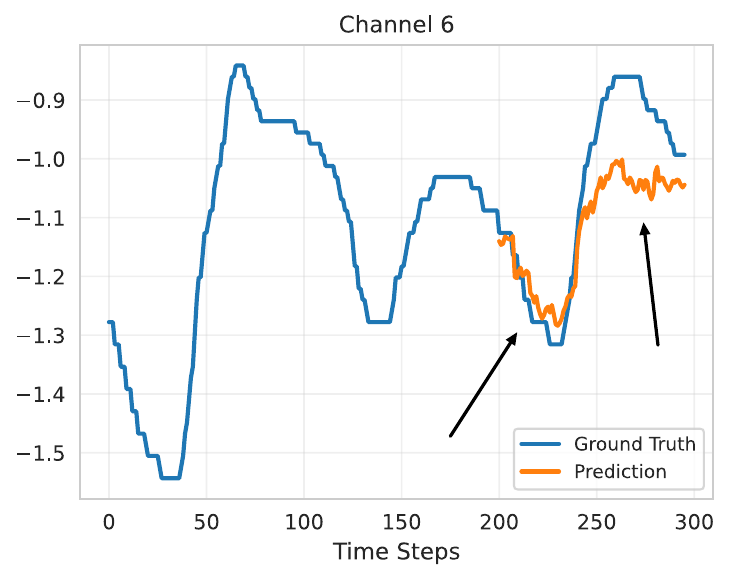}
            \caption{Standard MoE}
            \label{fig:show_moe_ettm2}
        \end{subfigure}
        \caption{Forecast showcases from \textbf{ETTm2} with a forecast horizon of 96. Blue curves are the ground truths, and orange curves are the model predictions. The curves before the model predictions are the input data.}
        \label{fig:show_ettm2}
    \end{figure}

\begin{figure}[!ht]
        \centering
        \begin{subfigure}{0.5\textwidth}
            \centering
            \includegraphics[width=0.82\textwidth]{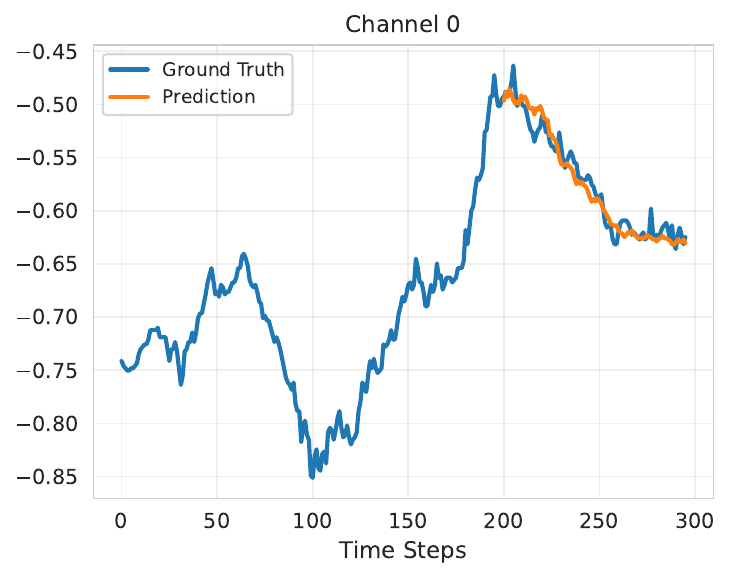}
            \caption{\model}
            \label{fig:show_segmoe_weather}
        \end{subfigure}%
        \begin{subfigure}{0.5\textwidth}
            \centering
            \includegraphics[width=0.82\textwidth]{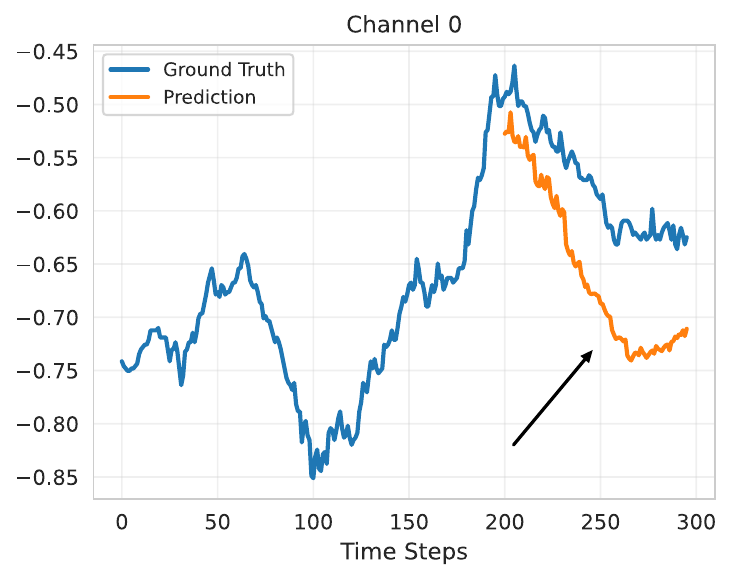}
            \caption{Standard MoE}
            \label{fig:show_moe_weather}
        \end{subfigure}
        \caption{Forecast showcases from \textbf{Weather} with a forecast horizon of 96. Blue curves are the ground truths, and orange curves are the model predictions. The curves before the model predictions are the input data.}
        \label{fig:show_weather}
    \end{figure}

\begin{figure}[!ht]
        \centering
        \begin{subfigure}{0.5\textwidth}
            \centering
            \includegraphics[width=0.82\textwidth]{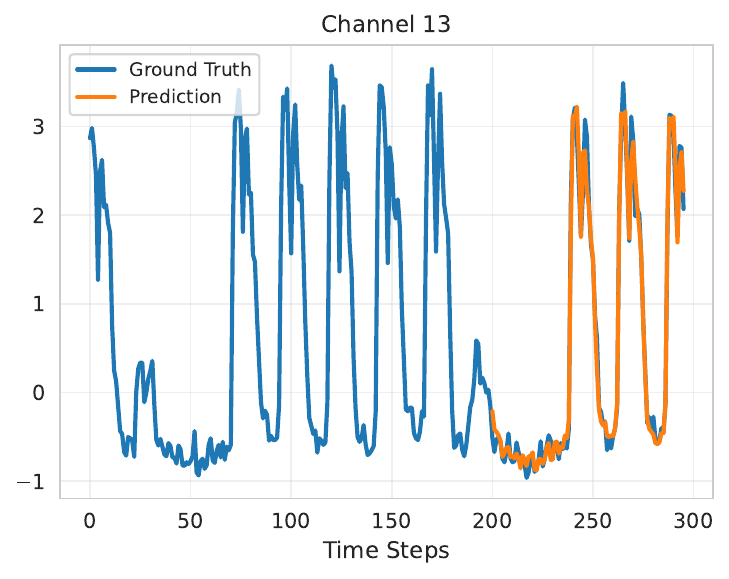}
            \caption{\model}
            \label{fig:show_segmoe_ecl}
        \end{subfigure}%
        \begin{subfigure}{0.5\textwidth}
            \centering
            \includegraphics[width=0.82\textwidth]{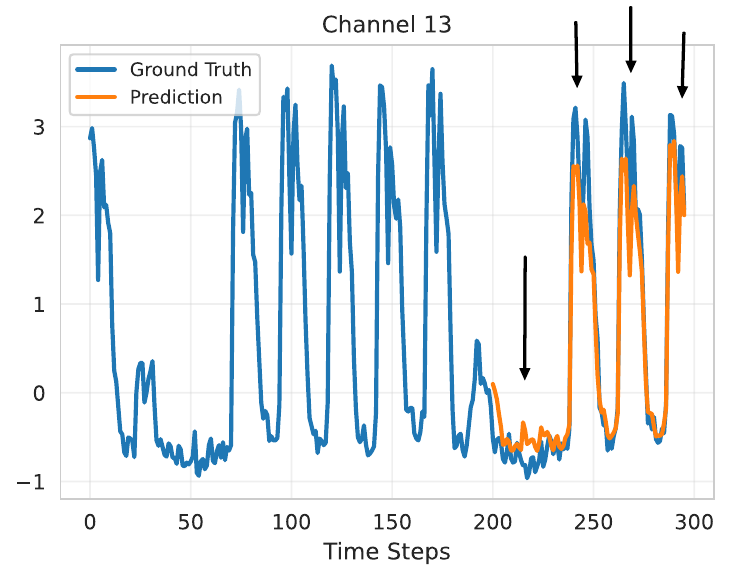}
            \caption{Standard MoE}
            \label{fig:show_moe_ecl}
        \end{subfigure}
        \caption{Forecast showcases from \textbf{ECL} with a forecast horizon of 96. Blue curves are the ground truths, and orange curves are the model predictions. The curves before the model predictions are the input data.}
        \label{fig:show_ecl}
    \end{figure}

\begin{figure}[!ht]
        \centering
        \begin{subfigure}{0.5\textwidth}
            \centering
            \includegraphics[width=0.82\textwidth]{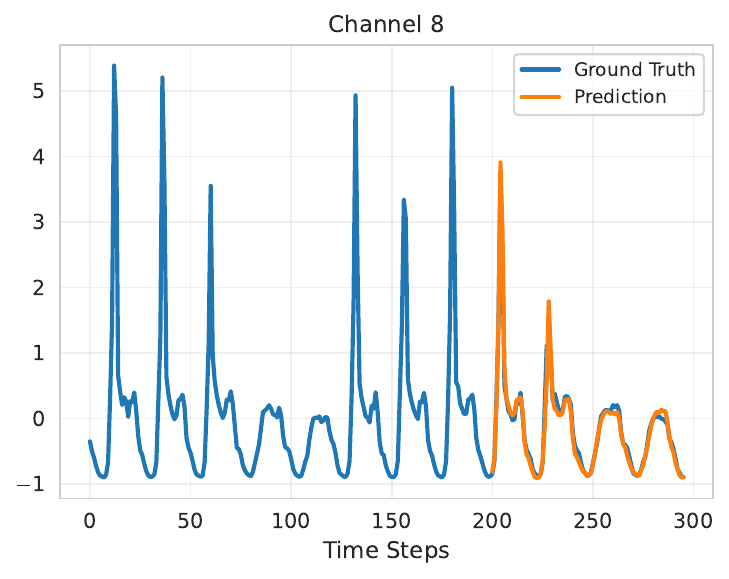}
            \caption{\model}
            \label{fig:show_segmoe_traffic}
        \end{subfigure}%
        \begin{subfigure}{0.5\textwidth}
            \centering
            \includegraphics[width=0.82\textwidth]{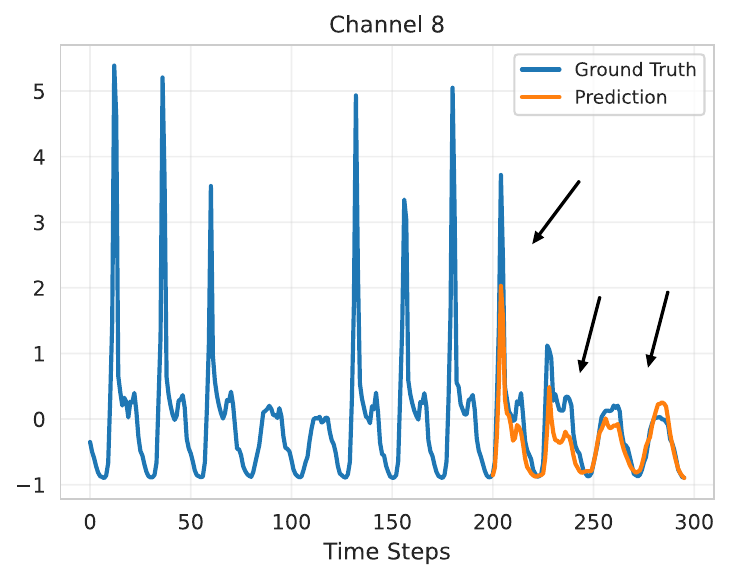}
            \caption{Standard MoE}
            \label{fig:show_moe_traffic}
        \end{subfigure}
        \caption{Forecast showcases from \textbf{Traffic} with a forecast horizon of 96. Blue curves are the ground truths, and orange curves are the model predictions. The curves before the model predictions are the input data.}
        \label{fig:show_traffic}
    \end{figure}

    Qualitative visualizations complement quantitative evaluations in long-term forecasting, revealing patterns that are not fully captured, such as phase shifts, amplitude attenuation, and delayed trend changes~\cite{shi2024time,nie2022time,wang2024timexer,wangtimemixer}. In this context, we visualize representative test-set forecasts from each benchmark dataset (i.e., ETTh1, ETTh2, ETTm1, ETTm2, Weather, ECL, and Traffic) using a fixed forecast horizon of $96$ time steps (Figures~\ref{fig:show_etth1}--\ref{fig:show_traffic}). Each figure contains two subfigures: one generated with \model and the other with a standard token-wise $\operatorname{MoE}$ baseline, both using a similar backbone and training protocol. For clarity, we plot the entire forecast horizon along with a short slice of the look-back context.
    
    To facilitate visual comparisons, we annotate representative regions with small black arrows highlighting where the two models diverge significantly; these regions typically correspond to transitions or peaks whose correct extrapolation benefits from segment-level processing.
    \model consistently yields tighter fits to the ground truth compared to token-wise $\operatorname{MoE}$.  
    As we can see, \model more accurately tracks local slope changes and oscillatory structures, and reduces common artifacts such as oversmoothing, lagged responses, or muted peaks.
    In general, the forecast illustrations align with the quantitative improvements reported in the main results (Tables~\ref{tab:ltsf_full} and \ref{tab:ltsf_additional}), where segment-wise routing yields forecasts that are not only lower in average error but also more temporally coherent.

\section{Limitations and Future Work}
\label{apdx:limitations_future}

    Initially designed for natural language processing, the Transformer architecture has demonstrated flexibility to be successfully applied to other domains, such as time series~\cite{nie2022time,shi2024time,liu2024timerxl}. However, natural language and time series data differ fundamentally. Time series are often generated by continuous, stochastic processes with irregular dynamics, whereas language exhibits a more deterministic grammatical and semantic structure. \model addresses this gap by design: it segments continuous time series into contiguous temporal segments, thereby encoding sequence continuity and recurring seasonality.

    Although \model demonstrates significant capabilities, specific directions warrant new challenges and opportunities. The multi-resolution segmentation enhances flexibility in modeling heterogeneous time patterns, but it also increases complexity. Each \model layer requires a chosen segment length, which introduces additional hyperparameters to tune. Manually selecting segment sizes for each layer can be demanding, especially for deep models. A natural future direction is to make segment resolution adaptive, enabling the model to learn optimal segment lengths. Furthermore, a \model layer operates at a fixed resolution, but we can imagine an $\operatorname{MoE}$ layer that supports multiple segment sizes by having parallel branches that patch the input at different granularities. Such extra flexibility would increase architectural and implementation complexity, since it requires dynamically partitioning the input and more sophisticated routing, but it could reduce manual tuning and improve robustness to varying patterns.

    We implemented non-overlapping segmentation for simplicity, but overlapping segmentation (sliding windows) is an alternative~\cite{nie2022time}. Incorporating overlapping or sliding segments into \model could help the model learn subtle transitions at segment boundaries. Overlapping windows introduce redundancy and extra computation. However, exploring sliding or partially overlapping segments is an interesting avenue that could improve coverage of boundary effects and continuity.

    Another promising extension is to diversify the expert architectures. In the current \model version, all experts share the same network architecture ($\operatorname{FFNs}$). Incorporating experts with different inductive biases could enable the model to capture a wider variety of temporal patterns~\cite{ortigossa2026mohets}. Such architectural diversity tends to improve specialization, since each expert can focus on patterns that suit its architecture. Finally, a major future direction is to scale \model-based models and pre-train them on large-scale time-series datasets to yield zero-shot forecasting performance~\cite{shi2024time,liu2024timerxl}, combining the benefits of segment-wise inductive bias and scalable sparse $\operatorname{MoE}$ to enable efficient large-scale training.

\end{document}